\pdfoutput=1
\documentclass[11pt]{article}

\PassOptionsToPackage{table}{xcolor}
\usepackage[review]{coling}

\usepackage{times}
\usepackage{latexsym}
\usepackage{amsmath}
\usepackage{algorithm}
\usepackage{algorithmic}
\usepackage{booktabs}
\usepackage{multirow}
\usepackage{tabularx}
\usepackage{longtable}
\usepackage{caption}
\usepackage[T1]{fontenc}
\usepackage[utf8]{inputenc}

\usepackage{microtype}

\usepackage{inconsolata}

\usepackage{graphicx}
\nolinenumbers % 禁用行号
\title{T3: A Novel Zero-shot Transfer Learning Framework Iteratively Training on an Assistant Task for a Target Task}

\author{
 \textbf{Xindi Tong\textsuperscript{1,2}},
 \textbf{Yujin Zhu\textsuperscript{1}},
 \textbf{Shijian Fan\textsuperscript{1}},
 \textbf{Liang Xu\textsuperscript{1}}
\\
\\
 \textsuperscript{1}Artificial Intelligence Innovation Center, \\ Yangtze Delta Region Institute of Tsinghua University \\
 \textsuperscript{2}Nanyang technological university\\
 \href{mailto:xinditong@ntu.edu}{to0001di@e.ntu.edu.sg}, \href{mailto:yujinzhu@tsinghua.edu}{zhuyujin@tsinghua-zj.edu.cn}, \href{mailto:shijianfan@tsinghua.edu}{fanshijian@tsinghua-zj.edu.cn},\\
 \textbf{Correspondence:} \href{mailto:liangxu@tsinghua.edu}{xlpaul@126.com}
}

\begin{document}
\maketitle
\begin{abstract}
Long text summarization, gradually being essential for efficiently processing large volumes of information, stays challenging for Large Language Models (LLMs) such as GPT and LLaMA families because of the insufficient open-sourced training datasets and the high requirement of contextual details dealing. To address the issue, we design a novel zero-shot transfer learning framework, abbreviated as T3, to iteratively training a baseline LLM on an assistant task for the target task, where the former should own richer data resources and share structural or semantic similarity with the latter. In practice, T3 is approached to deal with the long text summarization task by utilizing question answering as the assistant task, and further validated its effectiveness on the BBC summary, NarraSum, FairytaleQA, and NLQuAD datasets, with up to nearly 14\% improvement in ROUGE, 35\% improvement in BLEU, and 16\% improvement in Factscore compared to three baseline LLMs, demonstrating its potential for more assistant-target task combinations.
\end{abstract}

\section{Introduction}
In recent years, long text summarization has gradually played an important role in various real-world scenarios such as current affairs commentary, book summaries, news aggregation, and literature research, benefiting from the fast development of Large Language Models (LLMs) \cite{nallapati2016summarunnerrecurrentneuralnetwork, vaswani2023attentionneed,dubey2024LLaMA,islam2024gpt}. Unlike classical summarization tasks, which typically summarize texts of around 500 words or less into 100 or fewer words, long text summarization involves dealing with much longer texts. Clearly, as the original text length increases, the amount of contextual details and relevance required in the summary also grows, bringing new challenges to LLMs.

Considering solutions, we find that long text summarization could be categorized to a more general task, which shares two crucial issues: (1) This task lacks sufficient diverse open-source datasets to post-train or fine-tune LLMs, which might be due to the task itself being relatively new and special, or because the labeling details of the task make the construction of its datasets so complex and costly, whatever through manual annotation or algorithmic generation. (2) The output of this task demands a high level of contextual details and consistence. This makes it challenging for existing prompt engineering techniques \cite{schick2021fewshottextgenerationpatternexploiting, chakraborty2023abstractive}, most of which rely on task-oriented predefined rules to tune LLMs, thus probably losing insight into richer contextual information, as well as failing to capture similar semantic laws behind different tasks.

In this paper, we propose a novel zero-shot \textbf{T}ransfer learning framework iteratively training on an assistant \textbf{T}ask for a target \textbf{T}ask (abbreviated as T3). T3 fine-tunes the chosen baseline LLM through the assistant task that must meet two basic criteria: (1) It should have relatively abundant open-source datasets to address the first "insufficient dataset" issue. (2) It should exhibit similarity at the structural or semantic level to the target task to address the second "more details" issue mentioned above. Moreover, inspired by the ability of emergence from the new generation of LLMs to understand more complex semantics, an iterative optimization strategy is adopted to enable the used LLM to learn deeper from the assistant task and summarize useful rules (called as "experiences" in this paper) automatically by itself. Hereafter, the fine-tuned LLM is applied to the target task in a zero-shot manner. Further in practice, with long text summarization as the target task, we find that the Question Answering task (QA) is suitable as the assistant task because it has various open-source datasets, its used text is relatively long, and its question-answer pairs naturally contain richer entities and relationships. In addition, the Question Generation task (QG) is used during the training process of T3 to help LLMs better understand common and different contextual features between the assistant and the target tasks. Our contributions are concluded as follows:

\begin{itemize}
\item[$\bullet$] We propose a novel zero-shot transfer learning framework named T3. Different from most previous works, the baseline model in T3 only learns from the assistant task and then directly runs for the target task in zero-shot way. Further, T3 facilitates the transfer of knowledge from QA and QG tasks to the long text summarization task.
\item[$\bullet$] We adopt an iteration strategy in training process to make the used baseline model conclude useful rules automatically, with specially designed prompts and selected metrics, in order to ensure the high quality of the summaries generated during the test process in T3.
\item[$\bullet$] We demonstrate significant effectiveness of T3 on four datasets across different tasks by comparing with seven representative baseline LLMs.
\end{itemize}

\section{Related Work}
\subsection{Summarization}
Early automatic text summarization has begun with Luhn's keyword frequency method and Edmundson’s positional weighting in the 1950s. Machine learning approaches have emerged in the 1990s, such as Kupiec's Bayesian classifier. Later, deep learning methods like TextRank \cite{mihalcea-tarau-2004-textrank} advance the field. Recent research focuses on improving faithfulness in abstractive summarization, with PEGASUS \cite{zhang2020pegasuspretrainingextractedgapsentences} and methods like ECC \cite{zhang-etal-2022-improving-faithfulness}. Other works include contrastive learning \cite{chen-etal-2021-improving}, graph attention \cite{zhu-etal-2021-enhancing}, and few-shot approaches \cite{schick2021fewshottextgenerationpatternexploiting}. LLMs, like GPT, also contribute to faithful summarization \cite{zhang2023extractivesummarizationchatgptfaithful}.

Texts with more than 1,000 words could be considered as long text \cite{tuteja-gonzalez-jucla-2023-long}. Transformer-based models are efficient to handling of lengthy texts. Longformer \cite{beltagy2020longformerlongdocumenttransformer} uses a local attention mechanism to reduce computational complexity, enabling efficient processing of longer texts without losing global context. HyperGraph Transformers introduced by \cite{zhang2022hegel} tackle the complexity and context retention issues in Transformer-based models. Additionally, datasets like SQuALITY are crucial for advancing long-document summarization \cite{wang2022squality}. GPT could also applied for long text summarization which shows better performance than LLaMA \cite{fan2024evascoreevaluationlongformsummarization}.

\subsection{Question Answering}
Datasets like NQ \cite{kwiatkowski-etal-2019-natural} and SQuAD \cite{rajpurkar-etal-2016-squad, rajpurkar-etal-2018-know} have been foundational for the task. NQ focuses on real-world queries with Wikipedia annotations, while SQuAD emphasizes extractive QA. MS MARCO \cite{bajaj2016ms} and DuReader \cite{he-etal-2018-dureader} have broaden the QA field with free-form answers and non-English tasks. Recent works like \cite{eo-etal-2023-towards} enhance QA diversity via multimodal inputs and iterative generation. Transformer-based models like BERT \cite{devlin2019bertpretrainingdeepbidirectional}, fine-tuned on NQ and SQuAD, improve QA precision and recall. Further, \cite{eo-etal-2023-towards} distills complex information for summarization, aiding coherent QA generation. Here, QA is used as an assistant task for the proposed T3 on the summarization task.

\subsection{Question Generation}
Earlier work by \cite{du-etal-2017-learning} has highlighten the effectiveness of sequence-to-sequence models for QG in reading comprehension, while \cite{song-etal-2018-leveraging} improve QG accuracy using a multi-perspective context-matching model. \cite{chali-baghaee-2018-automatic} explore summarization for opinion-based QG to reduce repetition. \cite{lyu-etal-2021-improving} introduce an unsupervised approach, utilizing summaries to generate diverse questions with less lexical overlap. Frameworks like MQAG by \cite{manakul-etal-2023-mqag} ensure coherence in QG through summarization. Recently, \cite{choudhary-du-2024-qaevent} has proposed the QAEVENT paradigm, using summarization to represent document-level events as question-answer pairs. In this paper, QG is considered as a generation strategy in the training process of T3.

\subsection{Zero-shot Transfer Learning}
Recent advancements in zero-shot transfer learning have significantly improved the capabilities of LLMs across diverse tasks to solve problems such as data misalignment, label mismatches, and the limitations of pre-trained models without further fine-tuning \cite{kojima2023largelanguagemodelszeroshot, zhao-etal-2023-pre}. Furthermore, multilingual LLMs have demonstrated impressive zero-shot learning abilities in multimodal settings and cross-lingual instruction tuning, showing that LLMs trained on one language can generalize to others \cite{hu2023large}. ARL2 introduces an adaptive retriever learning technique that aligns with LLMs for better zero-shot generalization, particularly in tasks like QA \cite{zhang-etal-2024-arl2}. ZeroG applying GPT models focuses on zero-shot transfer for graph-based tasks, pioneering cross-dataset node classification by combining graph representations with LLM \cite{li2024zerog}. In this paper, a novel zero-shot transfer learning strategy using both assistant and target task is proposed.

\begin{figure*}[ht]
\centering
\includegraphics[width=\textwidth]{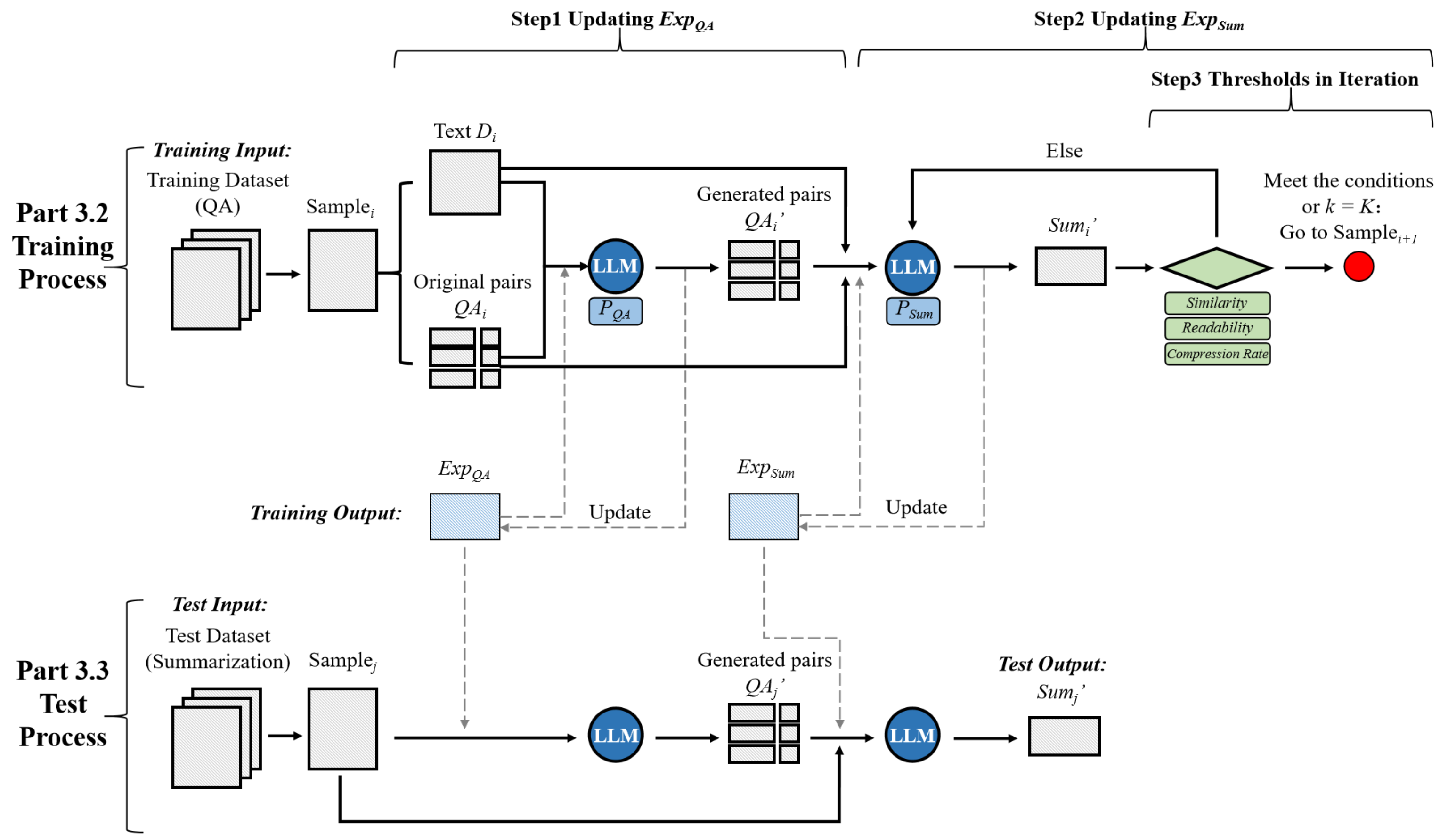}
\caption{Training and test process of T3 for summarization task.}
\label{fig:T3task}
\end{figure*}

\section{Methodology}
In this section, we first formulate the summarization task under T3, and dive into the detailed training and test processes. Then, we conclude the general form of the T3 workflow.

\subsection{Task Formulation}
Given the training dataset for Question-Answering task (QA), each sample from the dataset contains both the text $D_i$ and several labeled question-answer pairs $QA_i$ (where $i$ implies the $i$th sample), which form the text set $D = \{D_1, D_2, \ldots, D_n\}$ and the label set $QA = \{QA_1, QA_2, \ldots, QA_n\}$. Relatively, the question-answer pairs generated by LLM from $D_i$ are written as $QA_i'$. Given the test dataset for Summarization task (Sum), each sample from the dataset contains a text $D_j$, while LLM needs to generate the summary of the text $Sum_j'$. In training process, the prompts $P_{QA}$ and $P_{Sum}$ (see Appendix \ref{prompt_train_qa} and \ref{prompt_train_sum}) are used for LLM to generate the question-answer pairs $QA_i'$ and the summary $Sum_i'$, respectively. At the same time, LLM learns experiences from both tasks and records them in the form of several rules into $Exp_{QA}$ and $Exp_{Sum}$.

\subsection{Training Process}
The overview of T3 for summarization is illustrated in Figure \ref{fig:T3task}. The training process can be found at the top of the figure, while the test process is demonstrated at the bottom. Both inputs and outputs in these processes are written in bold in the figure. Furthermore, the detailed information of the training process is given in Algorithm \ref{alg:algorithm}, including the following key steps:

\label{training}
\begin{algorithm}[htb!]
\caption{T3 Training for Summarization}
\label{alg:algorithm}
\textbf{Input}: Text set $D = \{D_1, D_2, \ldots, D_n\}$ and label set $QA = \{QA_1, QA_2, \ldots, QA_n\}$ of the assistant QA task; Initialized experience $Exp_{QA}$ and $Exp_{Sum}$ for the QA and Summarization tasks, respectively.\\
\textbf{Parameter}: Predefined prompt $P_{QA}$ and $P_{Sum}$ for QA and Summary generation, respectively. Maximum iteration for each text $K$. Similarity $S$, readability $R$, and compression rate $C$ thresholds between the generated summary and the original text, respectively. The used baseline LLM $M$. \\
\textbf{Output}: Updated QA generation experience $Exp_{QA}$. Updated summary generation experience $Exp_{Sum}$

\begin{algorithmic}[1] %[1] enables line numbers
\STATE Initialize $Exp_{QA} \leftarrow \emptyset$, $Exp_{Sum} \leftarrow \emptyset$
\FOR{$i = 1$ to $n$}
    \STATE Generate QA pairs $QA_i'$ for $D_i$ using $Exp_{QA}$ based on $P_{QA}$ and model $M$.
    \STATE Learn from the generation process and update $Exp_{QA}$ using model $M$.
    \FOR{$k = 1$ to $K$}
        \STATE Generate summary $Sum_i'$ for $D_i$ using $Exp_{Sum}$ based on $P_{Sum}$, $QA_i'$, $QA_i$, and model $M$.
        \STATE Learn from the generation process and update $Exp_{Sum}$ using model $M$.
        \STATE Calculate the similarity between $Sum_i'$ and $D_i$ using cosine similarity $S_i$.
        \STATE Calculate the readability $R_i$ of $Sum_i'$ based on Equation \ref{eq1}.
        \STATE Calculate the compression rate $C_i$ of the generated summary based on Equation \ref{eq2}.
        \IF{$S_i > S$ and $R_i > R$ and $C_i < C$}
            \STATE break
        \ENDIF
    \ENDFOR
\ENDFOR
\end{algorithmic}
\end{algorithm}

\textbf{Step 1 Updating $Exp_{QA}$: } At first, both summary and QA generation experiences $Exp_{Sum}$ and $Exp_{QA}$ are initialized as $null$. Then, LLM generates new QA pairs $QA_i'$ for each text $D_i$ and updates $Exp_{QA}$ according to several rules concluded by itself through comparing $QA_i’$ with the ground truth $QA_i$. The prompt $P_{QA}$ is predefined to control all steps mentioned in this part.

\textbf{Step 2 Updating $Exp_{Sum}$: } In this part, LLM follows the predefined prompts $P_{Sum}$ and try to summarize the text $D_i$. As an prior-knowledge that might bring extra contextual information, both $QA_i$ and $QA_i’$ are also taken into account as the inputs. After the current summary $Sum_i'$ is generated, three thresholds are used to investigate whether the generated content is well enough. LLM would repeat the summary generation step and update its experience $Exp_{Sum}$ every iteration, till the stop conditions are fulfilled.

\textbf{Step 3 Thresholds Used in Iteration: } The iterative process mentioned above continues until the generated summary $Sum_i'$ simultaneously meets the required thresholds for similarity, readability \cite{flesch1948new}, and compression rate (or the number of iterations reach the preset maximum iteration times $K$). In details, readability $\text{R}$ is calculated as:

{\small
\begin{equation}
\label{eq1}
    \text{R} = 206.835 - 1.015 \times \left(\frac{TW}{TSE} \right) - 84.6 \times \left(\frac{TSY}{TW} \right)
\end{equation}
}

where, $TW$ is the total number of words in the given text, $TSE$ is the total number of sentences, and $TSY$ is the total number of syllables. $TW/ TSE $ calculates the average sentence length: a higher average sentence length makes the text harder to read, so it subtracts from the score. $TSY/TW$ calculates the average syllables per word: more syllables per word reduce the score because longer words tend to make the text more difficult to read. Moreover, the compression rate $\text{C}$ is calculated as:

{\small
\begin{equation}
\label{eq2}
    \text{C} = \text{length}(Summary)/\text{length}(Text)
\end{equation}
}

which is used to control the length of the generated summary.

\subsection{Test Process on Summarization Dataset}
According to the bottom of Figure \ref{fig:T3task}, during the test process, LLM would first use $Exp_{QA}$ learned from the training process to generated auxiliary question-answer pairs, and bring them with the updated $Exp_{Sum}$ to generate the final summary. Prompts in test process are given in Appendix \ref{prompttestsum} and a step-by-step use case is shown in Appendix \ref{step-by-step}.

\subsection{Test Process on QA Dataset}
In particular, as known from Figure \ref{fig:T3task}. We design QG, a more challenging text generation strategy for LLM, to replace the text understanding strategy of predicting answers for the questions from the given QA dataset during the training process, in order to better fit LLM to the summarization task. A by-product of this approach is that we can even run the QA dataset in test process to generate summaries for its own texts based on its QA pairs, and adopt appropriate metrics for effectiveness evaluation, as shown in Figure \ref{fig:testingQA}. Prompts in test process for QA dataset are given in Appendix \ref{prompttestqa}.

\begin{figure*}[ht!]
\centering
\includegraphics[width=\textwidth]{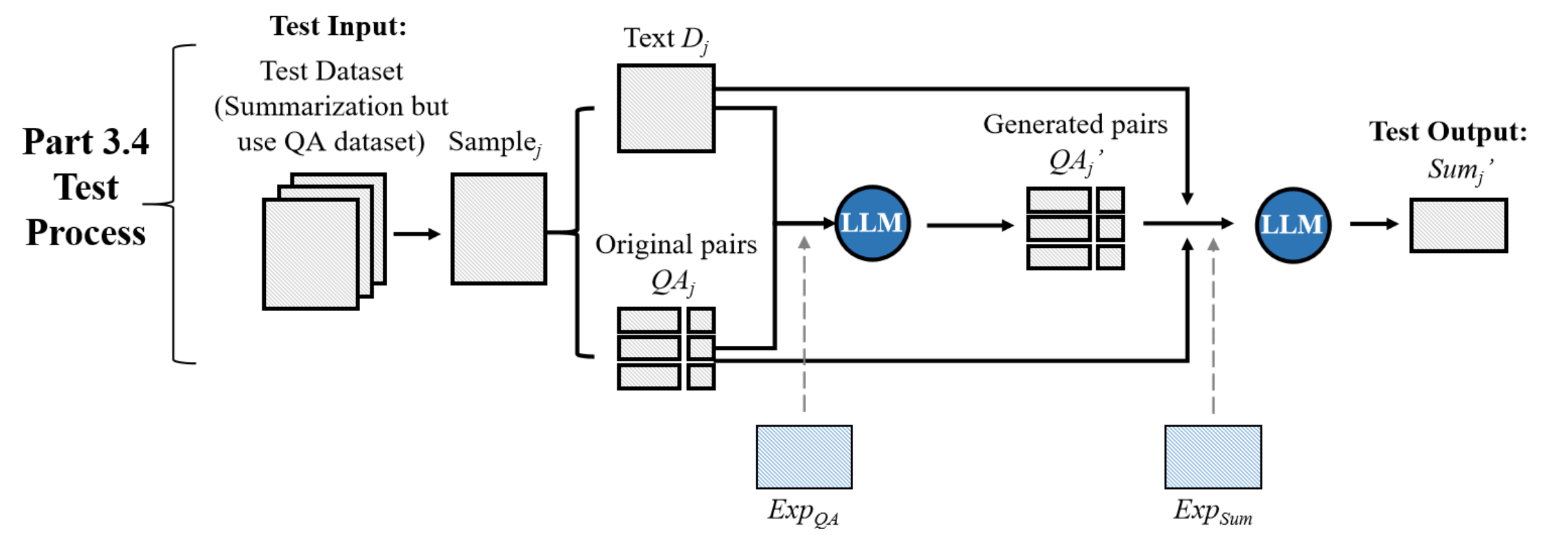}
\caption{Test process of T3 for summarization task on QA dataset.}
\label{fig:testingQA}
\end{figure*}

\subsection{T3 General Workflow}
The keypoint of T3 could be concluded that: in the whole training process, both the assistant task and the target task share the samples of the assistant task. At first, the used LLM \footnote{GPT-4o is recommended, while any other large models with sufficient parameter capacity to possess emergence could be considered.} acquires experience in handling the assistant task through the labeled data, which means a deeper understanding of the samples. Subsequently, LLM iteratively tries to accomplish the target task on these same samples and gain the experience for the target task by leveraging the previous learned contextual experiences, as well as being modulated by the predefined conditions. Finally, the model applies the experiences required for both the assistant and target tasks during the test process. Accordingly, a general workflow for task-agnostic T3 is shown in Figure \ref{fig:total_process} from  Appendix \ref{t3-general-fig}.

\section{Experiments} 
This section presents experimental details on four datasets, including two narrative dataset and two news datasets. We compare the quality of the summaries generated by the baseline models and our model using the generation algorithm from different aspects and perform an ablation study to analyze the influence of different components of T3.

\subsection{Settings}

\subsubsection{Datasets} Four datasets are used, including two summarization ones (BBC summary and NarraSum), and two QA ones (NLQuAD and FairytaleQA):

\textbf{BBC Summary} includes 2,225 BBC news from 2004-2005, used for categorization and summary research \cite{10.1145/1143844.1143892}.

\textbf{NarraSum} has 122K narrative synopses from movies and TV episodes, aimed at improving summarization research \cite{zhao-etal-2022-narrasum}.

\textbf{NLQuAD} contains 31,000 questions and answers derived from 13,000 BBC news, for non-factoid long QA, focusing on multi-sentence descriptive answers to enhance document-level language models and challenges models with long context and complex language understanding \cite{soleimani-etal-2021-NLQuAD}.

\textbf{FairytaleQA} includes 278 children's stories from Project Gutenberg, with questions and answers created by educational experts, aimed at assessing and training narrative comprehension for students (K-8) and models \cite{xu2022FairytaleQA}.

In experiments, 386 long articles from BBC summary, 260 long stories from FairytaleQA, 100 long stories from NarraSum, 501 long articles from NLQuAD are randomly picked up for evaluation. Each selected texts has a median word count exceeding 1,000. 

\subsubsection{Evaluation Metrics} To make more comprehensive evaluation, especially focusing on generated text quality and hallucinations alleviation, a range of classical and innovative metrics are used here:

\textbf{ROUGE} compares the overlap of n-grams, word sequences, and word pairs between the generated summary and reference summaries \cite{lin-2004-rouge}.

\textbf{BLEU} compares the overlap of n-grams between the generated text and reference translations, with higher scores indicating closer alignment with human translations and thus better quality \cite{papineni2002BLEU}.

\textbf{Factscore} We use FactSumm and ChatGPT for Factscore calculation. FactSumm uses named-entity recognition and relation extraction to extract facts from the source text and the generated summary to evaluate fact consistency by comparing extracted facts represented as triples (subject, relation, object) \cite{zhang-etal-2021-fine-grained}. ChatGPT evaluates factual inconsistency in summaries, outperforming previous methods with fewer hallucinations \cite{luo2023chatgptfactualinconsistencyevaluator}.

\subsubsection{Baseline Models}
Seven advanced LLMs from three most representative series are selected as the baselines. All of them are accessed through Application Programming Interface (API), which not only standardizes the experiment for reproducibility and consistency, but also saves computational resources.

\textbf{GPT Series} are developed by OpenAI and show strong language comprehension and generation capabilities for various natural language processing tasks \cite{achiam2023gpt}. Here, GPT-3.5-turbo, GPT-4 Turbo, and the newest lauched GPT-4o are adopted here.

\textbf{Claude Series} introduce advanced multimodal capabilities developed by Anthropic \cite{anthropic2023claude3}. Claude 3.5 Sonnet is the newest one of them, showing strong performance in graduate-level and text-based reasoning \cite{anthropic2023claude35sonnet}. Claude 3.5 Sonnet and Claude 3.0 Sonnet are chosen here.

\textbf{Gemini Series} surpasses Google's capabilities in understanding images, audio, video, and text \cite{team2023gemini}. Currently, Gemini-1.5-pro is the newest and best-performing one of them \cite{reid2024gemini}. Both Gemini-1.5-pro and Gemini-1.0-pro are used here.

\begin{table*}[ht!]
\centering

\begin{tabular}{lccccc}
\toprule
Models & ROUGE-1 & ROUGE-2 & ROUGE-L & BLEU  &  Factscore\\ \midrule

& \multicolumn{5}{c}{BBC summary} \\ \cmidrule(lr){2-6} 
&  w/o  \hspace{0.2cm}| \hspace{0.3cm}T3 &  w/o  \hspace{0.2cm}| \hspace{0.3cm}T3 &  w/o  \hspace{0.2cm}| \hspace{0.3cm}T3  &  w/o  \hspace{0.2cm}| \hspace{0.3cm}T3  &  w/o  \hspace{0.2cm}| \hspace{0.3cm}T3
                \\ \cmidrule(lr){2-6} 
\rowcolor{cyan!20} 
GPT-4o*                 & 40.19 | \textbf{41.71} & 18.14 | \textbf{21.45} & 36.97 | \textbf{39.44}   & 10.86 | \textbf{10.94}    & 63.64 | \textbf{70.12}   \\

GPT-4-turbo & \textbf{37.33} | 36.83 & \textbf{14.76} | 14.14 & \textbf{33.73} | 33.27 & \textbf{8.51} | 7.69  & 60.91 | \textbf{69.01} \\
GPT-3.5-turbo* & \textbf{39.36} | 37.88 & \textbf{17.75} | 16.95 & \textbf{36.04} | 34.78 & \textbf{10.50} | 8.58 &  \textbf{57.52} | 47.80 \\
\rowcolor{cyan!20} 
Gemini-1.5-pro* & 33.62 | \textbf{35.10} & 11.48 | \textbf{12.58} & 30.36 | \textbf{31.69} & 5.05 | \textbf{6.47} & 25.88 | \textbf{60.49} \\
Gemini-1.0-pro* & \textbf{34.90} | 33.58 & 12.80 | \textbf{13.92} & \textbf{31.75} | 31.17 & 5.99 | \textbf{7.22}  & 35.76 | \textbf{55.61} \\
\rowcolor{cyan!20} 
Claude-3.5-sonnet* & 39.08 | \textbf{43.54} & 16.59 | \textbf{21.13} & 35.88 | \textbf{40.36} & 9.70 | \textbf{13.00} & 63.43 | \textbf{65.83} \\
Claude-3.0-sonnet & \textbf{41.23} | 40.90 & \textbf{18.26} | 18.08 & \textbf{37.78} | 37.39 & \textbf{11.08} | 10.74 &  \textbf{60.02} | 50.43 \\
Average & 37.96 | \textbf{38.51} & 15.68 | \textbf{16.89} & 34.64 | \textbf{35.44} & 8.81 | \textbf{9.23} & 52.45 | \textbf{59.90} \\
 \midrule
& \multicolumn{5}{c}{NarraSum} \\ \cmidrule(lr){2-6} 
&  w/o  \hspace{0.2cm}| \hspace{0.3cm}T3 &  w/o  \hspace{0.2cm}| \hspace{0.3cm}T3 &  w/o  \hspace{0.2cm}| \hspace{0.3cm}T3  &  w/o  \hspace{0.2cm}| \hspace{0.3cm}T3  &  w/o  \hspace{0.2cm}| \hspace{0.3cm}T3
                \\ \cmidrule(lr){2-6} 
GPT-4o*                  &  \textbf{31.14} | 28.86 &  7.31 | \textbf{8.58} & \textbf{28.44} | 25.66    &  \textbf{2.68} | 2.55  & 97.75 | \textbf{99.69}   \\

GPT-4-turbo* & \textbf{29.14} | 27.27 & \textbf{7.72} | 7.42 & \textbf{25.44} | 24.45 & 3.36 | \textbf{3.60}  & \textbf{97.75} | 97.44 \\
GPT-3.5-turbo* & \textbf{28.60} | 25.28 & \textbf{7.52} | 5.88 & \textbf{26.50} | 23.00 & \textbf{2.79} | 2.15 &  95.20 | \textbf{95.91} \\
\rowcolor{cyan!20} 
Gemini-1.5-pro* & 25.99 | \textbf{26.95} & 5.21 | \textbf{5.39} & 25.00 | \textbf{25.32} & 1.75 | \textbf{1.99} & 93.26 | \textbf{93.57} \\
Gemini-1.0-pro & \textbf{28.97} | 28.25 & \textbf{6.26} | 5.74 & 25.45 | \textbf{26.36} & 1.65 | \textbf{2.09}  & \textbf{92.85} | 91.22 \\
\rowcolor{cyan!20} 
Claude-3.5-sonnet &  29.19 | \textbf{29.78} & 7.17 | \textbf{7.79} & 26.73 | \textbf{27.23} & 3.10 | \textbf{3.23} & 97.04 | \textbf{98.14} \\
Claude-3.0-sonnet* & \textbf{29.90} | 25.87 & \textbf{6.75} | 5.47 & \textbf{26.45} | 22.51 & \textbf{2.74} | 2.05 &  94.69 | \textbf{95.61} \\
Average  &\textbf{28.99} | 27.47 & \textbf{6.85} | \textbf{6.61} & \textbf{26.29} | 24.93 & \textbf{2.58} | 2.52 & 95.51 | \textbf{95.94} \\
 \midrule

& \multicolumn{5}{c}{NLQuAD} \\ \cmidrule(lr){2-6} 
&  w/o  \hspace{0.2cm}| \hspace{0.3cm}T3 &  w/o  \hspace{0.2cm}| \hspace{0.3cm}T3 &  w/o  \hspace{0.2cm}| \hspace{0.3cm}T3  &  w/o  \hspace{0.2cm}| \hspace{0.3cm}T3  &  w/o  \hspace{0.2cm}| \hspace{0.3cm}T3 
                \\ \cmidrule(lr){2-6} 
\rowcolor{cyan!20} 
GPT-4o* & 24.37 | \textbf{31.00} & 8.48 | \textbf{13.03} & 22.71 | \textbf{29.27} & 0.33 | \textbf{1.42} & 59.81 | \textbf{63.95} \\
GPT-4-turbo* & \textbf{24.19} | 23.66 & \textbf{7.39} | 7.36 & \textbf{22.16} | 21.62 & \textbf{0.44} | 0.36  & \textbf{53.33} | 49.16 \\

GPT-3.5-turbo* & 21.50 | \textbf{22.21} & 7.75 | \textbf{8.30} & 20.07 | \textbf{20.69} & 0.27 | \textbf{0.27} & \textbf{55.30} | 44.64 \\
\rowcolor{cyan!20} 
Gemini-1.5-pro* & 22.99 | \textbf{24.19} & 6.97 | \textbf{7.97} & 21.34 | \textbf{22.44} & 0.25 | \textbf{0.47} & 50.85 | \textbf{65.09} \\
\rowcolor{cyan!20} 
Gemini-1.0-pro* & 26.26 | \textbf{28.11} & 9.29 | \textbf{11.05} & 24.55 | \textbf{26.48} & 0.54 | \textbf{1.25}  & 64.57 | \textbf{71.64} \\
\rowcolor{cyan!20} 
Claude-3.5-sonnet* & 32.03 | \textbf{40.17} & 12.80 | \textbf{18.74} & 30.06 | \textbf{37.92} & 2.35 | \textbf{5.00}  & 65.71 | \textbf{82.76} \\
\rowcolor{cyan!20} 
Claude-3.0-sonnet* & 32.35 | \textbf{33.72} & 12.70 | \textbf{13.27} & 30.16 | \textbf{31.22} & 1.97 | \textbf{2.37}  & 58.89 | \textbf{61.69} \\
Average & 26.24 | \textbf{29.01} & 9.34 | \textbf{11.39} & 24.44 | \textbf{27.09} & 0.88 | \textbf{1.59} & 58.35 | \textbf{62.70} \\
 \midrule
& \multicolumn{5}{c}{FairytaleQA} \\ \cmidrule(lr){2-6} 
               &  w/o  \hspace{0.2cm}| \hspace{0.3cm}T3 &  w/o  \hspace{0.2cm}| \hspace{0.3cm}T3 &  w/o  \hspace{0.2cm}| \hspace{0.3cm}T3  &  w/o  \hspace{0.2cm}| \hspace{0.3cm}T3  &  w/o  \hspace{0.2cm}| \hspace{0.3cm}T3\\ \cmidrule(lr){2-6} 
\rowcolor{cyan!20} GPT-4o* & 13.34 | \textbf{14.86} & 3.15 | \textbf{3.72} & 12.34 | \textbf{13.78} & 0.04 | \textbf{0.07} & 98.55 | \textbf{98.92} \\
GPT-4-turbo* & \textbf{14.68} | 14.14 & \textbf{3.51} | 3.36 & \textbf{13.45} | 12.84 & 0.09 | \textbf{0.10} & 97.44 | \textbf{97.67} \\
GPT-3.5-turbo & 12.22 | \textbf{12.23} & 2.98 | \textbf{3.07} & \textbf{11.31} | 10.74 & 0.03 | \textbf{0.04} & \textbf{99.82} | 97.37 \\
\rowcolor{cyan!20} Gemini-1.5-pro* & 15.40 | \textbf{17.56} & 3.80 | \textbf{4.34} & 14.30 | \textbf{15.46} & 0.10 | \textbf{0.13} & 97.23 | \textbf{98.53} \\
Gemini-1.0-pro* & \textbf{14.14} | 13.43 & \textbf{3.25} | 2.43 & \textbf{14.22} | 13.39 & 0.03 | \textbf{0.14} & \textbf{97.84} | 94.24 \\
\rowcolor{cyan!20} Claude-3.5-sonnet* & 15.97 | \textbf{20.96} & 3.73 | \textbf{6.07} & 14.94 | \textbf{19.52} & 0.17 | \textbf{0.47} & 99.56 | \textbf{99.97} \\
\rowcolor{cyan!20} Claude-3.0-sonnet* & 17.20 | \textbf{18.91} & 4.08 | \textbf{4.90} & 15.81 | \textbf{17.35} & 0.17 | \textbf{0.33} & 97.47 | \textbf{98.92} \\
Average & 14.71 | \textbf{16.01} & 3.50 | \textbf{3.98} & 13.77 | \textbf{14.73} & 0.09 | \textbf{0.18} & \textbf{98.27} | 97.95 \\
\bottomrule

\end{tabular}

\caption{Experimental Results on BBC summary, NarraSum, NLQuAD and FairytaleQA datasets where each LLM without (left col) or with (right col) T3. }
\label{tab:my_label}
\end{table*}

% \begin{table*}[ht!]
% \centering
% \begin{tabular}{lcccc}
% \toprule
% Metrics & BBC summary & NarraSum & NLQuAD & FairytaleQA \\ \midrule
% ROUGE-1 & 0.3796 | 0.3851 & 0.2899 | 0.2747 & 0.2624 | 0.2901 & 0.1471 | 0.1601 \\
% ROUGE-2 & 0.1568 | 0.1689 & 0.0685 | 0.0661 & 0.0934 | 0.1139 & 0.0350 | 0.0398 \\
% ROUGE-L & 0.3464 | 0.3544 & 0.2629 | 0.2493 & 0.2444 | 0.2709 & 0.1377 | 0.1473 \\
% BLEU & 0.0881 | 0.0923 & 0.0258 | 0.0252 & 0.0088 | 0.0159 & 0.0009 | 0.0018 \\
% Factscore & 0.5245 | 0.5990 & 0.9551 | 0.9594 & 0.5835 | 0.6270 & 0.9827 | 0.9795 \\
% \bottomrule
% \end{tabular}
% \caption{Average Metrics for Each Dataset with and without T3}
% \label{tab:average_metrics}
% \end{table*}

\subsection{Implementation Details} 
The comparative experiment are designed and executed to validate both the feasibility and effectiveness of the proposed T3, by investigating the generated summarization results for all four datasets from all used LLMs with/without using T3.

The baseline summary of each text for each LLM is generated by using original text and the following prompt: "You are one helpful text assistant. Based on the input text, provide a summary in proper length. Aim to maximize Performance scores".\footnote{We add this to ensure fairness, since T3 use these words in its training process, according to Appendix \ref{prompt_train_sum}.} Correspondently, each LLM with T3 is guided by the experience and prompts derived from training process.

For BBC summary and NarraSum in test process, both the original texts from them, the QA generation experience $Exp_{QA}$ and the summary experience $Exp_{Sum}$ are contained into the input prompt, whose detailed prompts are given in Appendix \ref{expsumnew}, \ref{expsumna}, \ref{expqanew} and \ref{expqana} for News-style Datasets and Narratives-style Datasets respectively. 

For FairytaleQA and NLQuAD in test process, 10 texts and their corresponding QA pairs from each dataset are selected for training, while the remaining texts are used for testing. Both the original texts from the datasets, the QA generation experience $Exp_{QA}$ and the summary experience $Exp_{Sum}$ are contained into the input prompt, whose detailed prompts are given in Appendix \ref{expsumnew}, \ref{expsumna}, \ref{expqanew} and \ref{expqana} for News-style Datasets and Narratives-style Datasets respectively.

The higher scores from the used metrics are, indicating the better generation of summaries. It should be noted that the metric Factscore was calculated by GPT-4o for NarraSum and FairytaleQA, while calculated by FactSumm for BBC summary and NLQuAD, because it cannot extract facts from the story source text in practice so that does not perform well on narratives. 

For BBC summary and NarraSum, the provided summary was used as the reference, while for FairytaleQA and NLQuAD datasets, the original text was used as the reference. As mentioned in \textbf{Part 3.4}, T3 can also deal with QA datasets for summarization task. When the dataset does not provide reference summaries, the generated summary can be directly compared to the original text through certain automated metrics tools (e.g., ROUGE) \cite{louis-nenkova-2009-automatically}. According to \cite{saggion2013automatic}, the overlap of the generated summaries with the original text is important in assessing information coverage and the validity.

For models from GPT or Claude series, default values for temperature and other related hyper-parameters are applied. For Gemini models, we set 'HARASSMENT', 'HATE\_SPEECH', 'SEXUALLY\_EXPLICIT', 'DANGEROUS' in safety setting into 'block\_none'.

\subsection{Results}
Table \ref{tab:my_label} illustrates the performance of all used models with (right "T3" column) and without (left "w/o" column) T3 on all adopted datasets. The better one of the two values from each row of the two columns is marked in \textbf{bold}. Especially, for each row where all results from columns with T3 are better than those columns without T3, the row is colored \textbf{blue}. Models corresponding to results with significant differences between w/o and T3 are superscripted with *, where the detailed calculation process is given in Appendix \ref{significance}. From this table, we observe that: 

(1) LLMs using T3 outperforms themselves without T3 under most metrics of most datasets, which demonstates the feasibility and effectiveness of T3. Specially, the average improvement rates of GPT-4o, Gemini-1.5-pro, and Claude-3.5-sonnet using T3 compared to the baselines without T3 across different tasks are calculated and shown that these LLMs with T3 nearly improves the ROUGE score by 14\%, the BLEU score by 35\%, and the Factscore score by 16\%. Most of the comparison groups (82.14\%) showed significant differences with significance\_level=0.05 
 and null\_hypothesis that there is no significant difference between the means of the two groups.

(2) Of all the models in the four series, it is almost always the newest ones with T3 that achieve the most significant results comparing with itself without T3. For example, GPT-4o is 4\% and 5\% higher on Rouge-1 metrics compared to GPT-4  and GPT-3.5 on BBC Summary, respectively. These results could be due to the improvement of capabilities for latest LLMs on handling longer texts and more comprehensive tasks through understanding more complicated prompts and locating better details and events within lengthy texts \cite{anthropic2023claude35sonnet,islam2024gpt,reid2024gemini}. 

(3) Four used datasets are divided into news- and narrative-styles. For most models with/without T3, the generated summaries on narrative-style datasets NarraSum seems not as good as that of news-style ones BBC Summary. According to \cite{genette1980narrative, van2013news, barthes2016introduction}, the structure of news-style texts is compared with that of narrative-style texts, highlighting how news is easier to be summarized because of its concise layout and fixed pattern, while the narrative texts is more complexity, including chronological order, shifts in point of view, and narrative hierarchy, so that increasing the difficulty of summarization. 

(4) The improvement of LLMs with T3 seems more significant on NLQuAD and FairytaleQA, i.e., QA task-oriented datasets, which might be benefited from using the prior information of questions and answers pairs to enhance the contextual analyzing capability of T3. Moreover, incorporating QA information implies to improve summarization quality by enhancing text comprehension, focusing on key information, increasing coverage, ensuring coherence, and aligning summaries with user needs \cite{fabbri-etal-2022-qafacteval, siledar-etal-2024-product}.

\subsection{Ablation Study}
We further analyze the influence of different components of T3. 

Table \ref{tab:Table3} in Appendix \ref{abstudy} shows the experimental results for all used models after removing the summary experience component $Exp_{Sum}$ or the QA generation experience component $Exp_{QA}$, on all used datasets. It could be found that: (1) For models with less reasoning ability, longer inputs and complex instructions might diminish their performance mainly in their  $Exp_{Sum}$ component than their $Exp_{QA}$ component, while for more advanced models, both the two components could be utilized to enhance their overall effectiveness. (2) Since $Exp_{Sum}$ and $Exp_{QA}$ is run for the target task and the assistant task, respectively, it might be demonstrated the importance of the assistant task in the whole T3 workflow.

\section{Conclusion}
A novel T3 framework is proposed to address the challenge of long text summarization, using QA datasets to iteratively train baseline model at first, then to deal with summarization in a zero-shot way. Experiments on four datasets among seven representative baselines under five evaluation metrics validate the significant effectiveness of T3. Since T3 is model-agnostic and theoritically could be used for any suitable assistant task and target task pair, the future work will aim to explore the potential of it in a broader range of task combinations. 

\section*{Limitations}
When using the proposed T3 framework in real scenarios, the following limitations that need to be further optimized:

\textbf{Limited Baseline Models} Due to the costs and time constraints of API, we only test T3 on models from GPT, Claude, and Gemini series. More open-source models (such as LLaMA) and their private deployment situation will be considered in the future. 

\textbf{Limited Tasks} Few empirical guidelines for the selection of assistant tasks are given in this paper. In the future, the matching rules of assistant tasks and target tasks will be comprehensively explored in various task combinations.

\textbf{Limited Evaluation Metrics} As to evaluate generation results, BLEU and ROUGE focus on word overlap and neglect fluency, semantics, and factual accuracy, while FactSumm aims to consistency but struggles with abstract concepts. For summarization task, more suitable evaluation metrics should be discussed. 

\section*{Ethics Statement}

\textbf{Data Availability and Safety}
The training and test data used for summarization and QA in this paper are publicly available. While the original datasets were filtered, some content may still include sensitive material, such as reports on violent crimes or incidents. Furthermore, when utilizing the Gemini model service, the safety mode is set to 'block\_none,' which may result in outputs containing violent or explicit content.

\textbf{Usage of Large Language Models}
Models like GPT, Gemini, and Claude are utilized to generate QA results, refine prompt descriptions, and produce text summaries for input texts in summarization tasks. The generated text is used exclusively for experimental and analytical purposes, and the findings are presented in the relevant sections.

% \section*{Acknowledgments}
% % This work is supported by the Artificial Intelligence Innovation Center, Yangtze Delta Region Institute of Tsinghua University. 
% We thank the anonymous reviewers for their helpful feedback.


\bibliography{custom}

\begin{thebibliography}{56}
\providecommand{\natexlab}[1]{#1}

\bibitem[{Achiam et~al.(2023)Achiam, Adler, Agarwal, Ahmad, Akkaya, Aleman, Almeida, Altenschmidt, Altman, Anadkat et~al.}]{achiam2023gpt}
Josh Achiam, Steven Adler, Sandhini Agarwal, Lama Ahmad, Ilge Akkaya, Florencia~Leoni Aleman, Diogo Almeida, Janko Altenschmidt, Sam Altman, Shyamal Anadkat, et~al. 2023.
\newblock Gpt-4 technical report.
\newblock \emph{arXiv preprint arXiv:2303.08774}.

\bibitem[{Anthropic(2023{\natexlab{a}})}]{anthropic2023claude35sonnet}
Anthropic. 2023{\natexlab{a}}.
\newblock Claude 3.5 sonnet.
\newblock \url{https://www.anthropic.com/news/claude-3-5-sonnet}.
\newblock Accessed: 2024-09-02.

\bibitem[{Anthropic(2023{\natexlab{b}})}]{anthropic2023claude3}
Anthropic. 2023{\natexlab{b}}.
\newblock Introducing claude 3.0.
\newblock \url{https://www.anthropic.com}.
\newblock Accessed: 2023-09-02.

\bibitem[{Bajaj et~al.(2016)Bajaj, Campos, Craswell, Deng, Gao, Liu, Majumder, McNamara, Mitra, Nguyen et~al.}]{bajaj2016ms}
Payal Bajaj, Daniel Campos, Nick Craswell, Li~Deng, Jianfeng Gao, Xiaodong Liu, Rangan Majumder, Andrew McNamara, Bhaskar Mitra, Tri Nguyen, et~al. 2016.
\newblock Ms marco: A human generated machine reading comprehension dataset.
\newblock \emph{arXiv preprint arXiv:1611.09268}.

\bibitem[{Barthes(2016)}]{barthes2016introduction}
Roland Barthes. 2016.
\newblock Introduction to the structural analysis of narratives.
\newblock In \emph{Myths and Mythologies}, pages 290--307. Routledge.

\bibitem[{Beltagy et~al.(2020)Beltagy, Peters, and Cohan}]{beltagy2020longformerlongdocumenttransformer}
Iz~Beltagy, Matthew~E. Peters, and Arman Cohan. 2020.
\newblock \href {https://arxiv.org/abs/2004.05150} {Longformer: The long-document transformer}.
\newblock \emph{Preprint}, arXiv:2004.05150.

\bibitem[{Chakraborty and Pakray(2023)}]{chakraborty2023abstractive}
Shayak Chakraborty and Partha Pakray. 2023.
\newblock Abstractive summarization evaluation for prompt engineering.
\newblock In \emph{International Visual Informatics Conference}, pages 629--640. Springer.

\bibitem[{Chali and Baghaee(2018)}]{chali-baghaee-2018-automatic}
Yllias Chali and Tina Baghaee. 2018.
\newblock \href {https://doi.org/10.18653/v1/W18-6518} {Automatic opinion question generation}.
\newblock In \emph{Proceedings of the 11th International Conference on Natural Language Generation}, pages 152--158, Tilburg University, The Netherlands. Association for Computational Linguistics.

\bibitem[{Chen et~al.(2021)Chen, Zhang, Sone, and Roth}]{chen-etal-2021-improving}
Sihao Chen, Fan Zhang, Kazoo Sone, and Dan Roth. 2021.
\newblock \href {https://doi.org/10.18653/v1/2021.naacl-main.475} {Improving faithfulness in abstractive summarization with contrast candidate generation and selection}.
\newblock In \emph{Proceedings of the 2021 Conference of the North American Chapter of the Association for Computational Linguistics: Human Language Technologies}, pages 5935--5941, Online. Association for Computational Linguistics.

\bibitem[{Choudhary and Du(2024)}]{choudhary-du-2024-qaevent}
Milind Choudhary and Xinya Du. 2024.
\newblock \href {https://aclanthology.org/2024.findings-eacl.126} {{QAEVENT}: Event extraction as question-answer pairs generation}.
\newblock In \emph{Findings of the Association for Computational Linguistics: EACL 2024}, pages 1860--1873, St. Julian{'}s, Malta. Association for Computational Linguistics.

\bibitem[{Devlin et~al.(2019)Devlin, Chang, Lee, and Toutanova}]{devlin2019bertpretrainingdeepbidirectional}
Jacob Devlin, Ming-Wei Chang, Kenton Lee, and Kristina Toutanova. 2019.
\newblock \href {https://arxiv.org/abs/1810.04805} {Bert: Pre-training of deep bidirectional transformers for language understanding}.
\newblock \emph{Preprint}, arXiv:1810.04805.

\bibitem[{Du et~al.(2017)Du, Shao, and Cardie}]{du-etal-2017-learning}
Xinya Du, Junru Shao, and Claire Cardie. 2017.
\newblock \href {https://doi.org/10.18653/v1/P17-1123} {Learning to ask: Neural question generation for reading comprehension}.
\newblock In \emph{Proceedings of the 55th Annual Meeting of the Association for Computational Linguistics (Volume 1: Long Papers)}, pages 1342--1352, Vancouver, Canada. Association for Computational Linguistics.

\bibitem[{Dubey et~al.(2024)Dubey, Jauhri, Pandey, Kadian, Al-Dahle, Letman, Mathur, Schelten, Yang, Fan et~al.}]{dubey2024LLaMA}
Abhimanyu Dubey, Abhinav Jauhri, Abhinav Pandey, Abhishek Kadian, Ahmad Al-Dahle, Aiesha Letman, Akhil Mathur, Alan Schelten, Amy Yang, Angela Fan, et~al. 2024.
\newblock The llama 3 herd of models.
\newblock \emph{arXiv preprint arXiv:2407.21783}.

\bibitem[{Eo et~al.(2023)Eo, Moon, Kim, Hur, Kim, Lee, Chun, Park, and Lim}]{eo-etal-2023-towards}
Sugyeong Eo, Hyeonseok Moon, Jinsung Kim, Yuna Hur, Jeongwook Kim, SongEun Lee, Changwoo Chun, Sungsoo Park, and Heuiseok Lim. 2023.
\newblock \href {https://doi.org/10.18653/v1/2023.findings-acl.380} {Towards diverse and effective question-answer pair generation from children storybooks}.
\newblock In \emph{Findings of the Association for Computational Linguistics: ACL 2023}, pages 6100--6115, Toronto, Canada. Association for Computational Linguistics.

\bibitem[{Fabbri et~al.(2022)Fabbri, Wu, Liu, and Xiong}]{fabbri-etal-2022-qafacteval}
Alexander Fabbri, Chien-Sheng Wu, Wenhao Liu, and Caiming Xiong. 2022.
\newblock \href {https://doi.org/10.18653/v1/2022.naacl-main.187} {{QAF}act{E}val: Improved {QA}-based factual consistency evaluation for summarization}.
\newblock In \emph{Proceedings of the 2022 Conference of the North American Chapter of the Association for Computational Linguistics: Human Language Technologies}, pages 2587--2601, Seattle, United States. Association for Computational Linguistics.

\bibitem[{Fan et~al.(2024)Fan, Zhong, Wang, Wu, and Zhou}]{fan2024evascoreevaluationlongformsummarization}
Yuchen Fan, Xin Zhong, Chengsi Wang, Gaoche Wu, and Bowen Zhou. 2024.
\newblock \href {https://arxiv.org/abs/2407.04969} {Eva-score: Evaluation of long-form summarization on informativeness through extraction and validation}.
\newblock \emph{Preprint}, arXiv:2407.04969.

\bibitem[{Flesch(1948)}]{flesch1948new}
Rudolph Flesch. 1948.
\newblock A new readability yardstick.
\newblock \emph{Journal of applied psychology}, 32(3):221.

\bibitem[{Genette(1980)}]{genette1980narrative}
G{\'e}rard Genette. 1980.
\newblock Narrative discourse: An essay in method.
\newblock \emph{Cornell UP}.

\bibitem[{Greene and Cunningham(2006)}]{10.1145/1143844.1143892}
Derek Greene and P\'{a}draig Cunningham. 2006.
\newblock \href {https://doi.org/10.1145/1143844.1143892} {Practical solutions to the problem of diagonal dominance in kernel document clustering}.
\newblock In \emph{Proceedings of the 23rd International Conference on Machine Learning}, ICML '06, page 377–384, New York, NY, USA. Association for Computing Machinery.

\bibitem[{He et~al.(2018)He, Liu, Liu, Lyu, Zhao, Xiao, Liu, Wang, Wu, She, Liu, Wu, and Wang}]{he-etal-2018-dureader}
Wei He, Kai Liu, Jing Liu, Yajuan Lyu, Shiqi Zhao, Xinyan Xiao, Yuan Liu, Yizhong Wang, Hua Wu, Qiaoqiao She, Xuan Liu, Tian Wu, and Haifeng Wang. 2018.
\newblock \href {https://doi.org/10.18653/v1/W18-2605} {{D}u{R}eader: a {C}hinese machine reading comprehension dataset from real-world applications}.
\newblock In \emph{Proceedings of the Workshop on Machine Reading for Question Answering}, pages 37--46, Melbourne, Australia. Association for Computational Linguistics.

\bibitem[{Hu et~al.(2023)Hu, Yao, Wang, Wang, Pan, Chen, Yu, Wu, Zhao, Zhang et~al.}]{hu2023large}
Jinyi Hu, Yuan Yao, Chongyi Wang, Shan Wang, Yinxu Pan, Qianyu Chen, Tianyu Yu, Hanghao Wu, Yue Zhao, Haoye Zhang, et~al. 2023.
\newblock Large multilingual models pivot zero-shot multimodal learning across languages.
\newblock \emph{arXiv preprint arXiv:2308.12038}.

\bibitem[{Islam and Moushi(2024)}]{islam2024gpt}
Raisa Islam and Owana~Marzia Moushi. 2024.
\newblock Gpt-4o: The cutting-edge advancement in multimodal llm.
\newblock \emph{Authorea Preprints}.

\bibitem[{Kojima et~al.(2023)Kojima, Gu, Reid, Matsuo, and Iwasawa}]{kojima2023largelanguagemodelszeroshot}
Takeshi Kojima, Shixiang~Shane Gu, Machel Reid, Yutaka Matsuo, and Yusuke Iwasawa. 2023.
\newblock \href {https://arxiv.org/abs/2205.11916} {Large language models are zero-shot reasoners}.
\newblock \emph{Preprint}, arXiv:2205.11916.

\bibitem[{Kwiatkowski et~al.(2019)Kwiatkowski, Palomaki, Redfield, Collins, Parikh, Alberti, Epstein, Polosukhin, Devlin, Lee, Toutanova, Jones, Kelcey, Chang, Dai, Uszkoreit, Le, and Petrov}]{kwiatkowski-etal-2019-natural}
Tom Kwiatkowski, Jennimaria Palomaki, Olivia Redfield, Michael Collins, Ankur Parikh, Chris Alberti, Danielle Epstein, Illia Polosukhin, Jacob Devlin, Kenton Lee, Kristina Toutanova, Llion Jones, Matthew Kelcey, Ming-Wei Chang, Andrew~M. Dai, Jakob Uszkoreit, Quoc Le, and Slav Petrov. 2019.
\newblock \href {https://doi.org/10.1162/tacl_a_00276} {Natural questions: A benchmark for question answering research}.
\newblock \emph{Transactions of the Association for Computational Linguistics}, 7:452--466.

\bibitem[{Li et~al.(2024)Li, Wang, Li, Yu, and Li}]{li2024zerog}
Yuhan Li, Peisong Wang, Zhixun Li, Jeffrey~Xu Yu, and Jia Li. 2024.
\newblock Zerog: Investigating cross-dataset zero-shot transferability in graphs.
\newblock In \emph{Proceedings of the 30th ACM SIGKDD Conference on Knowledge Discovery and Data Mining}, pages 1725--1735.

\bibitem[{Lin(2004)}]{lin-2004-rouge}
Chin-Yew Lin. 2004.
\newblock \href {https://aclanthology.org/W04-1013} {{ROUGE}: A package for automatic evaluation of summaries}.
\newblock In \emph{Text Summarization Branches Out}, pages 74--81, Barcelona, Spain. Association for Computational Linguistics.

\bibitem[{Louis and Nenkova(2009)}]{louis-nenkova-2009-automatically}
Annie Louis and Ani Nenkova. 2009.
\newblock \href {https://aclanthology.org/D09-1032} {Automatically evaluating content selection in summarization without human models}.
\newblock In \emph{Proceedings of the 2009 Conference on Empirical Methods in Natural Language Processing}, pages 306--314, Singapore. Association for Computational Linguistics.

\bibitem[{Luo et~al.(2023)Luo, Xie, and Ananiadou}]{luo2023chatgptfactualinconsistencyevaluator}
Zheheng Luo, Qianqian Xie, and Sophia Ananiadou. 2023.
\newblock \href {https://arxiv.org/abs/2303.15621} {Chatgpt as a factual inconsistency evaluator for text summarization}.
\newblock \emph{Preprint}, arXiv:2303.15621.

\bibitem[{Lyu et~al.(2021)Lyu, Shang, Graham, Foster, Jiang, and Liu}]{lyu-etal-2021-improving}
Chenyang Lyu, Lifeng Shang, Yvette Graham, Jennifer Foster, Xin Jiang, and Qun Liu. 2021.
\newblock \href {https://doi.org/10.18653/v1/2021.emnlp-main.340} {Improving unsupervised question answering via summarization-informed question generation}.
\newblock In \emph{Proceedings of the 2021 Conference on Empirical Methods in Natural Language Processing}, pages 4134--4148, Online and Punta Cana, Dominican Republic. Association for Computational Linguistics.

\bibitem[{Manakul et~al.(2023)Manakul, Liusie, and Gales}]{manakul-etal-2023-mqag}
Potsawee Manakul, Adian Liusie, and Mark Gales. 2023.
\newblock \href {https://doi.org/10.18653/v1/2023.ijcnlp-main.4} {{MQAG}: Multiple-choice question answering and generation for assessing information consistency in summarization}.
\newblock In \emph{Proceedings of the 13th International Joint Conference on Natural Language Processing and the 3rd Conference of the Asia-Pacific Chapter of the Association for Computational Linguistics (Volume 1: Long Papers)}, pages 39--53, Nusa Dua, Bali. Association for Computational Linguistics.

\bibitem[{Mihalcea and Tarau(2004)}]{mihalcea-tarau-2004-textrank}
Rada Mihalcea and Paul Tarau. 2004.
\newblock \href {https://aclanthology.org/W04-3252} {{T}ext{R}ank: Bringing order into text}.
\newblock In \emph{Proceedings of the 2004 Conference on Empirical Methods in Natural Language Processing}, pages 404--411, Barcelona, Spain. Association for Computational Linguistics.

\bibitem[{Nallapati et~al.(2016)Nallapati, Zhai, and Zhou}]{nallapati2016summarunnerrecurrentneuralnetwork}
Ramesh Nallapati, Feifei Zhai, and Bowen Zhou. 2016.
\newblock \href {https://arxiv.org/abs/1611.04230} {Summarunner: A recurrent neural network based sequence model for extractive summarization of documents}.
\newblock \emph{Preprint}, arXiv:1611.04230.

\bibitem[{Papineni et~al.(2002)Papineni, Roukos, Ward, and Zhu}]{papineni2002BLEU}
Kishore Papineni, Salim Roukos, Todd Ward, and Wei-Jing Zhu. 2002.
\newblock Bleu: a method for automatic evaluation of machine translation.
\newblock In \emph{Proceedings of the 40th annual meeting of the Association for Computational Linguistics}, pages 311--318.

\bibitem[{Rajpurkar et~al.(2018)Rajpurkar, Jia, and Liang}]{rajpurkar-etal-2018-know}
Pranav Rajpurkar, Robin Jia, and Percy Liang. 2018.
\newblock \href {https://doi.org/10.18653/v1/P18-2124} {Know what you don{'}t know: Unanswerable questions for {SQ}u{AD}}.
\newblock In \emph{Proceedings of the 56th Annual Meeting of the Association for Computational Linguistics (Volume 2: Short Papers)}, pages 784--789, Melbourne, Australia. Association for Computational Linguistics.

\bibitem[{Rajpurkar et~al.(2016)Rajpurkar, Zhang, Lopyrev, and Liang}]{rajpurkar-etal-2016-squad}
Pranav Rajpurkar, Jian Zhang, Konstantin Lopyrev, and Percy Liang. 2016.
\newblock \href {https://doi.org/10.18653/v1/D16-1264} {{SQ}u{AD}: 100,000+ questions for machine comprehension of text}.
\newblock In \emph{Proceedings of the 2016 Conference on Empirical Methods in Natural Language Processing}, pages 2383--2392, Austin, Texas. Association for Computational Linguistics.

\bibitem[{Reid et~al.(2024)Reid, Savinov, Teplyashin, Lepikhin, Lillicrap, Alayrac, Soricut, Lazaridou, Firat, Schrittwieser et~al.}]{reid2024gemini}
Machel Reid, Nikolay Savinov, Denis Teplyashin, Dmitry Lepikhin, Timothy Lillicrap, Jean-baptiste Alayrac, Radu Soricut, Angeliki Lazaridou, Orhan Firat, Julian Schrittwieser, et~al. 2024.
\newblock Gemini 1.5: Unlocking multimodal understanding across millions of tokens of context.
\newblock \emph{arXiv preprint arXiv:2403.05530}.

\bibitem[{Saggion and Poibeau(2013)}]{saggion2013automatic}
Horacio Saggion and Thierry Poibeau. 2013.
\newblock Automatic text summarization: Past, present and future.
\newblock \emph{Multi-source, multilingual information extraction and summarization}, pages 3--21.

\bibitem[{Schick and Schütze(2021)}]{schick2021fewshottextgenerationpatternexploiting}
Timo Schick and Hinrich Schütze. 2021.
\newblock \href {https://arxiv.org/abs/2012.11926} {Few-shot text generation with pattern-exploiting training}.
\newblock \emph{Preprint}, arXiv:2012.11926.

\bibitem[{Siledar et~al.(2024)Siledar, Rangaraju, Muddu, Banerjee, Patil, Singh, Chelliah, Garera, Nath, and Bhattacharyya}]{siledar-etal-2024-product}
Tejpalsingh Siledar, Rupasai Rangaraju, Sankara Muddu, Suman Banerjee, Amey Patil, Sudhanshu Singh, Muthusamy Chelliah, Nikesh Garera, Swaprava Nath, and Pushpak Bhattacharyya. 2024.
\newblock \href {https://doi.org/10.18653/v1/2024.findings-naacl.150} {Product description and {QA} assisted self-supervised opinion summarization}.
\newblock In \emph{Findings of the Association for Computational Linguistics: NAACL 2024}, pages 2315--2332, Mexico City, Mexico. Association for Computational Linguistics.

\bibitem[{Soleimani et~al.(2021)Soleimani, Monz, and Worring}]{soleimani-etal-2021-NLQuAD}
Amir Soleimani, Christof Monz, and Marcel Worring. 2021.
\newblock \href {https://doi.org/10.18653/v1/2021.eacl-main.106} {{NLQ}u{AD}: A non-factoid long question answering data set}.
\newblock In \emph{Proceedings of the 16th Conference of the European Chapter of the Association for Computational Linguistics: Main Volume}, pages 1245--1255, Online. Association for Computational Linguistics.

\bibitem[{Song et~al.(2018)Song, Wang, Hamza, Zhang, and Gildea}]{song-etal-2018-leveraging}
Linfeng Song, Zhiguo Wang, Wael Hamza, Yue Zhang, and Daniel Gildea. 2018.
\newblock \href {https://doi.org/10.18653/v1/N18-2090} {Leveraging context information for natural question generation}.
\newblock In \emph{Proceedings of the 2018 Conference of the North {A}merican Chapter of the Association for Computational Linguistics: Human Language Technologies, Volume 2 (Short Papers)}, pages 569--574, New Orleans, Louisiana. Association for Computational Linguistics.

\bibitem[{Team et~al.(2023)Team, Anil, Borgeaud, Wu, Alayrac, Yu, Soricut, Schalkwyk, Dai, Hauth et~al.}]{team2023gemini}
Gemini Team, Rohan Anil, Sebastian Borgeaud, Yonghui Wu, Jean-Baptiste Alayrac, Jiahui Yu, Radu Soricut, Johan Schalkwyk, Andrew~M Dai, Anja Hauth, et~al. 2023.
\newblock Gemini: a family of highly capable multimodal models.
\newblock \emph{arXiv preprint arXiv:2312.11805}.

\bibitem[{Tuteja and Gonz{\'a}lez~Jucl{\`a}(2023)}]{tuteja-gonzalez-jucla-2023-long}
Mohit Tuteja and Daniel Gonz{\'a}lez~Jucl{\`a}. 2023.
\newblock \href {https://doi.org/10.18653/v1/2023.nllp-1.3} {Long text classification using transformers with paragraph selection strategies}.
\newblock In \emph{Proceedings of the Natural Legal Language Processing Workshop 2023}, pages 17--24, Singapore. Association for Computational Linguistics.

\bibitem[{Van~Dijk(2013)}]{van2013news}
Teun~A Van~Dijk. 2013.
\newblock \emph{News as discourse}.
\newblock Routledge.

\bibitem[{Vaswani et~al.(2023)Vaswani, Shazeer, Parmar, Uszkoreit, Jones, Gomez, Kaiser, and Polosukhin}]{vaswani2023attentionneed}
Ashish Vaswani, Noam Shazeer, Niki Parmar, Jakob Uszkoreit, Llion Jones, Aidan~N. Gomez, Lukasz Kaiser, and Illia Polosukhin. 2023.
\newblock \href {https://arxiv.org/abs/1706.03762} {Attention is all you need}.
\newblock \emph{Preprint}, arXiv:1706.03762.

\bibitem[{Wang et~al.(2022)Wang, Pang, Chen, Phang, and Bowman}]{wang2022squality}
Alex Wang, Richard~Yuanzhe Pang, Angelica Chen, Jason Phang, and Samuel~R. Bowman. 2022.
\newblock S{Q}u{ALITY}: Building a long-document summarization dataset the hard way.
\newblock \emph{arXiv preprint 2205.11465}.

\bibitem[{Xu et~al.(2022)Xu, Wang, Yu, Ritchie, Yao, Wu, Zhang, Li, Bradford, Sun, Hoang, Sang, Hou, Ma, Yang, Peng, Yu, and Warschauer}]{xu2022FairytaleQA}
Ying Xu, Dakuo Wang, Mo~Yu, Daniel Ritchie, Bingsheng Yao, Tongshuang Wu, Zheng Zhang, Toby Jia-Jun Li, Nora Bradford, Branda Sun, Tran~Bao Hoang, Yisi Sang, Yufang Hou, Xiaojuan Ma, Diyi Yang, Nanyun Peng, Zhou Yu, and Mark Warschauer. 2022.
\newblock Fantastic questions and where to find them: Fairytale{QA} -- an authentic dataset for narrative comprehension.
\newblock Association for Computational Linguistics.

\bibitem[{Zhang et~al.(2022{\natexlab{a}})Zhang, Liu, and Zhang}]{zhang2022hegel}
Haopeng Zhang, Xiao Liu, and Jiawei Zhang. 2022{\natexlab{a}}.
\newblock Hegel: Hypergraph transformer for long document summarization.
\newblock \emph{arXiv preprint arXiv:2210.04126}.

\bibitem[{Zhang et~al.(2023)Zhang, Liu, and Zhang}]{zhang2023extractivesummarizationchatgptfaithful}
Haopeng Zhang, Xiao Liu, and Jiawei Zhang. 2023.
\newblock \href {https://arxiv.org/abs/2304.04193} {Extractive summarization via chatgpt for faithful summary generation}.
\newblock \emph{Preprint}, arXiv:2304.04193.

\bibitem[{Zhang et~al.(2022{\natexlab{b}})Zhang, Yavuz, Kryscinski, Hashimoto, and Zhou}]{zhang-etal-2022-improving-faithfulness}
Haopeng Zhang, Semih Yavuz, Wojciech Kryscinski, Kazuma Hashimoto, and Yingbo Zhou. 2022{\natexlab{b}}.
\newblock \href {https://doi.org/10.18653/v1/2022.findings-naacl.40} {Improving the faithfulness of abstractive summarization via entity coverage control}.
\newblock In \emph{Findings of the Association for Computational Linguistics: NAACL 2022}, pages 528--535, Seattle, United States. Association for Computational Linguistics.

\bibitem[{Zhang et~al.(2020)Zhang, Zhao, Saleh, and Liu}]{zhang2020pegasuspretrainingextractedgapsentences}
Jingqing Zhang, Yao Zhao, Mohammad Saleh, and Peter~J. Liu. 2020.
\newblock \href {https://arxiv.org/abs/1912.08777} {Pegasus: Pre-training with extracted gap-sentences for abstractive summarization}.
\newblock \emph{Preprint}, arXiv:1912.08777.

\bibitem[{Zhang et~al.(2024)Zhang, Yu, Wang, and Zhang}]{zhang-etal-2024-arl2}
LingXi Zhang, Yue Yu, Kuan Wang, and Chao Zhang. 2024.
\newblock \href {https://aclanthology.org/2024.acl-long.203} {{ARL}2: Aligning retrievers with black-box large language models via self-guided adaptive relevance labeling}.
\newblock In \emph{Proceedings of the 62nd Annual Meeting of the Association for Computational Linguistics (Volume 1: Long Papers)}, pages 3708--3719, Bangkok, Thailand. Association for Computational Linguistics.

\bibitem[{Zhang et~al.(2021)Zhang, Niu, and Wei}]{zhang-etal-2021-fine-grained}
Sen Zhang, Jianwei Niu, and Chuyuan Wei. 2021.
\newblock \href {https://doi.org/10.18653/v1/2021.emnlp-main.9} {Fine-grained factual consistency assessment for abstractive summarization models}.
\newblock In \emph{Proceedings of the 2021 Conference on Empirical Methods in Natural Language Processing}, pages 107--116, Online and Punta Cana, Dominican Republic. Association for Computational Linguistics.

\bibitem[{Zhao et~al.(2022)Zhao, Brahman, Song, Yao, Yu, and Chaturvedi}]{zhao-etal-2022-narrasum}
Chao Zhao, Faeze Brahman, Kaiqiang Song, Wenlin Yao, Dian Yu, and Snigdha Chaturvedi. 2022.
\newblock \href {https://doi.org/10.18653/v1/2022.findings-emnlp.14} {{N}arra{S}um: A large-scale dataset for abstractive narrative summarization}.
\newblock In \emph{Findings of the Association for Computational Linguistics: EMNLP 2022}, pages 182--197, Abu Dhabi, United Arab Emirates. Association for Computational Linguistics.

\bibitem[{Zhao et~al.(2023)Zhao, Ouyang, Yu, Wu, and Li}]{zhao-etal-2023-pre}
Xuandong Zhao, Siqi Ouyang, Zhiguo Yu, Ming Wu, and Lei Li. 2023.
\newblock \href {https://doi.org/10.18653/v1/2023.acl-long.869} {Pre-trained language models can be fully zero-shot learners}.
\newblock In \emph{Proceedings of the 61st Annual Meeting of the Association for Computational Linguistics (Volume 1: Long Papers)}, pages 15590--15606, Toronto, Canada. Association for Computational Linguistics.

\bibitem[{Zhu et~al.(2021)Zhu, Hinthorn, Xu, Zeng, Zeng, Huang, and Jiang}]{zhu-etal-2021-enhancing}
Chenguang Zhu, William Hinthorn, Ruochen Xu, Qingkai Zeng, Michael Zeng, Xuedong Huang, and Meng Jiang. 2021.
\newblock \href {https://doi.org/10.18653/v1/2021.naacl-main.58} {Enhancing factual consistency of abstractive summarization}.
\newblock In \emph{Proceedings of the 2021 Conference of the North American Chapter of the Association for Computational Linguistics: Human Language Technologies}, pages 718--733, Online. Association for Computational Linguistics.

\end{thebibliography}

\newpage 
\appendix
\onecolumn
\section{Appendix}

\subsection{General workflow of task-agnostic T3}
The figure below shows the general task-agnostic workflow of T3, i.e., the overall workflow of T3 when both the assistant and target tasks are not identified. The workflow facilitates the matching of different assistant and target tasks in practice.
\label{t3-general-fig}
\begin{figure*}[ht!]
\centering
\includegraphics[width=0.8\textwidth]{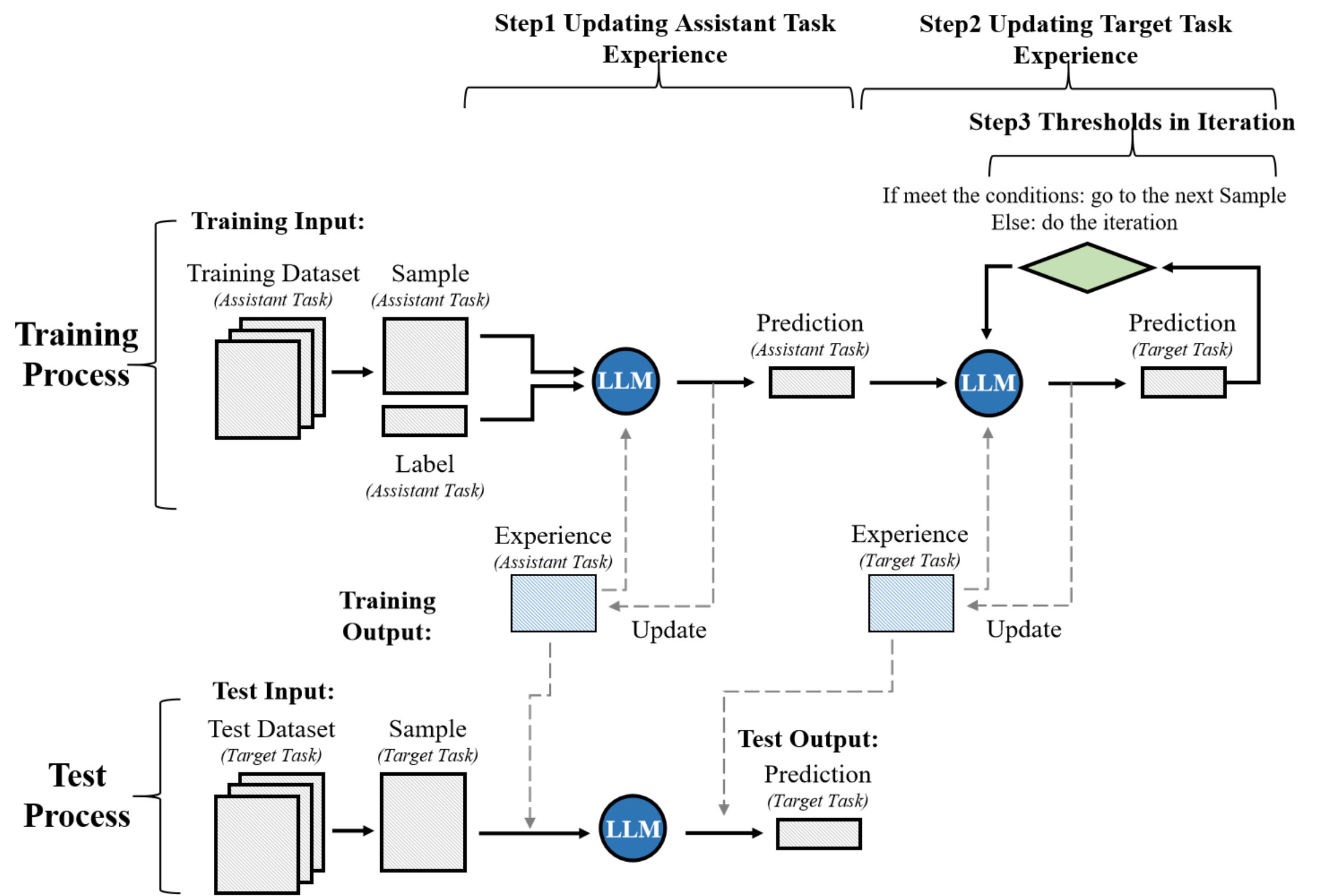}
\caption{General workflow of task-agnostic T3}
\label{fig:total_process}
\end{figure*}

\subsection{Prompt for QA Generation and Experience Updating in Training Process}
\noindent
\begin{tabular}{|p{0.95\linewidth}|}
  % \hline
  % \textbf{Contents} \\ 
  \hline
You are a helpful text assistant skilled at learning new experiences. First, please help us with the following:
\\
Based on the input text, provided questions, and QA pair generation experience, generate questions and corresponding answers that reflect all subjects and key points of the input article. The format of the generated QA pairs should match that of the provided question pairs.
\\
Update the learning experience for generating better QA pairs during the question generation process by comparing the generated questions to the original question pairs based on the original text or other methods. Consider all key points needed to create good QA pairs, and ensure the learning experience is general. DO NOT INCLUDE DETAILED EXAMPLES BUT PROVIDE RULES AS DETAILED AS POSSIBLE. DO NOT EXCEED 600 WORDS.
\\
The format of the output is as follows:\\
\{
  "Generated\_QA\_pairs": \{
    "1": { "Question": "xxxx", "Answer": "xxxx" },
    "2": { "Question": "xxxx", "Answer": "xxxx" }
  \},
  "QA\_generation\_experience": "Answer point by point text in one line"
\}\\
\hline
\end{tabular}
\label{prompt_train_qa}
\newpage

\subsection{Prompt for Summarization Generation and Experience Updating in Training Process}
\noindent
\begin{tabular}{|p{0.95\linewidth}|}
  % \hline
  % \textbf{Contents} \\ 
  \hline
You are a helpful text assistant. Please help us with the following:
\\
1. Based on the input text, provided questions, generated QA pairs, and summary generation experience, generate a summary that corresponds to the original text, reflecting all subjects and key points of the input article at an appropriate length.
\\
2. Update the summary generation experience with insights on how to create a better summary by first using the provided questions, then the generated summary, and all provided questions (both generated and original), and finally incorporating what you've learned during question generation in the question generation experience.
\\
Here is the target for the generated summary:\\
THE GENERATED SUMMARY SHOULD HAVE HIGHER PERFORMANCE SCORES COMPARED TO THE DIRECTLY GENERATED ONES. The summary should include essential terms, phrases, and sentences that closely mirror the original text. Aim to maximize the overlap with the reference summary by closely following the wording, sentence structure, and style of the original text. Ensure that the summary is concise but comprehensive, avoiding unnecessary details or deviations. Also, consider the fluency of the summary.
\\
Consider all key points to make a good summary, and make the learning experience general (e.g., avoid losing details in the story, maintain sentence coherence, and list some points to improve accuracy and fluency). Additionally, you should list other rules that might be important, with each point as detailed as possible.
\\
You may also consider methods to ensure a better summary, such as voting.
\\
Answer in one line and point by point. DO NOT EXCEED 600 WORDS.
\\
The format of the output is as follows:\\
\{ "Summary": "text in one line",  \\
"Summary\_generation\_experience": "Answer point by point text in one line" \}.\\

You must follow this format.\\
\hline
\end{tabular}
\label{prompt_train_sum}

\subsection{Prompt in Test Process on Summarization Dataset}
\noindent
\begin{tabular}{|p{0.95\linewidth}|}
 % \hline
 % \textbf{Contents} \\ 
  \hline
You are a helpful text assistant. Please help us with the following:
\\
Based on the input text, and with the help of QA generation rules and summary rules, here are the steps to create the summary:
\\
First, generate QA pairs based on the generation rules.  
Then, based on the input text, provided questions, generated QA pairs, and the summary generation experience, generate a summary of the original text at the appropriate length.
\\
Article: [Article]  \\
Summary Generation Experience: [Summary generation experience]  \\
QA Generation Experience: [QA generation experience]\\
  \hline
\end{tabular}
\label{prompttestsum}
\vspace{1em}

\subsection{Prompt in Test Process on QA Dataset}
\noindent
\begin{tabular}{|p{0.95\linewidth}|}
  % \hline
  % \textbf{Contents} \\ 
  \hline
You are a helpful text assistant. Please help us with the following:
\\
Based on the input text and the questions given, and with the help of QA generation rules and summary rules, here are the steps to create the summary:
\\
First, generate QA pairs based on the generation rules.  
Then, based on the input text, provided questions, generated QA pairs, and the summary generation experience, generate a summary of the original text at the appropriate length.
\\
Article: [Article]  \\
Question Pair: [Question Pair with answer]  \\
Summary Generation Experience: [Summary generation experience]  \\
QA Generation Experience: [QA generation experience] \\
  \hline
\end{tabular}
\label{prompttestqa}
\vspace{1em}
\newpage

\subsection{Summary Experience for News-style Datasets}
\noindent
\begin{tabular}{|p{0.95\linewidth}|}
  % \hline
  % \textbf{Contents} \\ 
  \hline
1. Ensure all key points, such as the formation, evolution, and current status of Boko Haram, as well as contrasting views on security, are included.  \\
2. Maintain the original wording and sentence structure for precision and recall.  \\
3. Avoid unnecessary details, focusing on core information for conciseness.  \\
4. Ensure logical flow and fluency in the summary.  \\
5. Use specific terms and phrases from the text to maximize overlap with the reference summary.  \\
6. Consider using a voting system to select the best summary from multiple drafts.  \\
7. Verify that the summary reflects the original text accurately without speculative information.  \\
8. Address distinct key points or subjects in the summary for comprehensive coverage.  \\
9. Compare generated summary with original text to ensure all critical aspects are covered.  \\
10. Ensure summary are directly supported by the text, avoiding speculative or inferred information.\\
  \hline
\end{tabular}
\label{expsumnew}
\vspace{1em}

\subsection{Summary Experience for Narratives-style Datasets}
\noindent
\begin{tabular}{|p{0.95\linewidth}|}
  % \hline
  % \textbf{Contents} \\ 
  \hline
1. Ensure all key points and characters from the text are covered to avoid missing crucial details; use details to enhance emotional impact, such as incorporating symbolic details.  \\
2. Maintain the original sequence of events to reflect the narrative flow accurately.  \\
3. Use language and phrases closely mirroring the original text for higher precision and recall.  \\
4. Keep the summary concise yet comprehensive, avoiding unnecessary details while ensuring all significant events are included.  \\
5. Focus on fluency, ensuring the summary reads smoothly and logically. Preserve the logical flow and connection between events for coherence.  \\
6. Review and refine the summary multiple times to maximize overlap with the reference summary.  \\
7. Pay attention to the connection of sentences to maintain coherence.  \\
8. Include essential terms and phrases to ensure high precision and recall.  \\
9. Emphasize themes and symbolism: Don’t just recount the plot; also highlight the underlying themes or symbolic meanings.  \\
10. Closely follow the wording, sentence structure, and style of the original text.  \\
11. Highlight the emotional tones and themes present in the original text. This adds depth to the summary and helps convey the underlying messages or morals of the story. It’s important not just to recount events but also to capture the feelings and ideas these events evoke.  \\
12. Ensure that no important details contributing to the story's moral or lesson are lost. These details are often critical to understanding the overall message of the text, so they should be preserved to maintain the integrity of the original story.  \\
13. Consider using a voting method to select the best summary from multiple versions, such as generating three versions and choosing the best one.\\
  \hline
\end{tabular}
\label{expsumna}
\vspace{1em}

\subsection{QA Generation Experience for News-style Datasets}
\noindent
\begin{tabular}{|p{0.95\linewidth}|}
  % \hline
  % \textbf{Contents} \\ 
  \hline
1. Ensure each question targets a distinct key point or subject from the original text.  \\
2. Questions should be specific to elicit detailed answers that reflect core information.  \\
3. Avoid broad or vague questions to prevent ambiguous answers.  \\
4. Compare generated questions with original ones to ensure all critical aspects are covered, and questions are clear and concise.  \\
5. Answers should be directly supported by the text and should avoid speculative or inferred information. This approach ensures comprehensive and accurate QA pairs.\\
  \hline
\end{tabular}
\label{expqanew}
\vspace{1em}

\subsection{QA Generation Experience for Narratives-style Datasets}
\noindent
\begin{tabular}{|p{0.95\linewidth}|}
  % \hline
  % \textbf{Contents} \\ 
  \hline
1. Ensure each question addresses a unique key point or detail to avoid redundancy. Cover all significant events, characters, and details for comprehensive coverage.\\
2. Keep questions clear and specific to avoid ambiguity and ensure they can be answered directly from the text.\\
3. Include a mix of factual, inferential, and interpretative questions to engage different levels of understanding.\\
4. Review the text multiple times to ensure no key points are missed and that questions flow logically.\\
5. Pay attention to character motivations, actions, and consequences to generate deeper questions.\\
6. Consider the context and sequence of events to create questions reflecting the narrative's progression.\\
7. Ensure answers are concise and directly derived from the text for accuracy.\\
8. Compare generated questions with existing ones to identify gaps or overlaps and refine accordingly.\\
9. Aim to create questions that encourage critical thinking and deeper engagement with the text.\\
  \hline
\end{tabular}
\label{expqana}
\vspace{1em}

\subsection{Step-by-step example}
This section provides a detailed use case description, which helps readers understand the operation flow of each step more intuitively by analyzing the application of T3 step by step.

\subsubsection{Original Text}
In this example, we apply one sample from the dataset NarraSum, a story containing more than 1700 words, as our summarization target. 

\noindent
\begin{tabular}{|p{0.95\linewidth}|}
\hline
         "The opening scene depicts Anita \"Needy\" Lesnicky (Amanda Seyfriend), a violent inmate in a correctional facility. She is notorious for being a 'kicker', and is now reflecting during lockdown on past events. Needy is a nerdy, substantially unpopular, \"plain Jane\" type student from Devil's Kettle, Minnesota, who has been best friends with Jennifer Check (Megan Fox), a beautiful, popular cheerleader, since their childhood. ...... The end credits show Low Shoulder checking into a hotel post-concert and candidly indulging in drugs and alcohol in their hotel room. Eventually, the band is savagely murdered by a blurred figure. We then see police crime scene photos of the band members' corpses strewn around the hotel room, Nikolai having been stabbed in the chest with the sacrificial knife. The final image of the movie shows security camera footage of an evil-looking Needy walking down the hotel hallway past a gathering of female band groupies."\\
         \hline
\end{tabular}

\subsubsection{Generate QA Pairs}
Using the sample and the QA generation experience in \ref{expqana} as input, the baseline model creates several question-answer pairs to capture key plot points and character details. Here are some examples:

\noindent
\begin{tabular}{|p{0.95\linewidth}|}
\hline

\textbf{Question}: "What is Needy's situation at the beginning of the movie?"\\
\textbf{Answer}: "She is an inmate in a correctional facility, known for her violent behavior as a 'kicker'."
\\

\textbf{Question}: "Who are the main characters and their relationship?"\\
\textbf{Answer}: "Needy Lesnicky, an unpopular student, and Jennifer Check, a popular cheerleader, have been best friends since childhood.\\

\textbf{Question}: "What happens to the band Low Shoulder at the end?"\\
\textbf{Answer}: "The band is murdered by a figure, later revealed as Needy, after a concert while indulging in drugs and alcohol.\\

\hline
\end{tabular}

\subsubsection{Final Summary}
Using the generated QA pairs and the summary experience in \ref{expsumna} as input, the baseline model generates a summary of the story as follows:\\

\vspace{1em}
\noindent
\begin{tabular}{|p{0.95\linewidth}|}
\hline

The story revolves around Anita \"Needy\" Lesnicky, a violent inmate reflecting on past events. Needy, a nerdy student from Devil's Kettle, Minnesota, was best friends with Jennifer Check, a popular cheerleader. One night, they attend a concert by indie rock band Low Shoulder at a local bar. The bar catches fire, and Jennifer leaves with the band despite Needy's protests. Later, Jennifer appears at Needy's house, injured and behaving strangely. The next day, Jennifer acts normal at school but soon begins killing and feeding on male students. Needy becomes suspicious and investigates, discovering that Low Shoulder had attempted to sacrifice Jennifer in a Satanic ritual for fame and success. However, because Jennifer wasn't a virgin, the ritual backfired, turning her into a succubus.Needy tries to warn her boyfriend Chip about Jennifer's true nature, but he doesn't believe her. At the Spring Formal, Jennifer attacks Chip, leading to a confrontation between Needy and Jennifer. Chip dies in Needy's arms after trying to save her. Consumed by revenge, Needy breaks into Jennifer's house and kills her. Needy is institutionalized for Jennifer's murder but develops supernatural abilities from being bitten during their fight. She escapes the institution and sets out to seek revenge on Low Shoulder. The story concludes with security footage showing Needy walking down a hotel hallway, implying she has murdered the band members. The narrative explores themes of friendship, betrayal, and the supernatural, while also critiquing the music industry and teenage social dynamics. It blends elements of horror, dark comedy, and coming-of-age drama to create a unique and compelling story.
\\
\hline
\end{tabular}
\vspace{1em}

This final summary keeps the plot concise while covering essential character arcs, the supernatural element, and the revenge ending. As shown in this example, by using QA to capture plot intricacies, the proposed T3 framework ensures that the final summary is both comprehensive and aligned with the main themes of the original content.
\label{step-by-step}

\subsection{Ablation study}
\label{abstudy}
\begin{longtable}{|l|c|c|c|c|c|c|}
\hline
Models & Strategy & ROUGE-1 & ROUGE-2 & ROUGE-L & BLEU  & Factscore \\ 
\hline
\endfirsthead
\hline
Models & & ROUGE-1 & ROUGE-2 & ROUGE-L & BLEU & Factscore \\ 
\hline
\endhead
\hline
\multicolumn{7}{|r|}{Continued on the next page...} \\
\hline
\endfoot
\hline
\endlastfoot

\multirow{16}{*}{GPT-3.5-turbo} & & \multicolumn{5}{|c|}{NLQuAD} \\ \cline{3-7}
& Full T3                 & 22.21 & 8.30 & 20.69 & 0.27  & 47.80 \\
& w/o summary & 31.68 & 15.57 & 30.24 & 5.31 & 46.36 \\
& w/o qa      & 26.01 & 9.59 & 23.91 & 0.83 & 45.35 \\
 \cline{2-7}
 & &\multicolumn{5}{|c|}{BBC Summary} \\  \cline{3-7} 
& Full T3                 & 37.88 & 16.95 & 34.78 & 8.58  & 47.80 \\
& w/o summary  & 40.62 & 22.74 & 37.05 & 12.66 & 55.49 \\
& w/o qa     & 36.00 & 15.07 & 32.39 & 8.14 & 46.08\\
 \cline{2-7}   
 & &\multicolumn{5}{|c|}{FairyTaleQA} \\  \cline{3-7} 
& Full T3                 & 12.23 & 3.07 & 10.74 & 0.04 &  97.37 \\
& w/o summary  & 15.11 & 3.56 & 13.81 & 0.12 & 98.34\\
& w/o qa     & 12.45 & 2.83 & 10.63 & 0.03 & 97.27\\
 \cline{2-7}  
 & &\multicolumn{5}{|c|}{NarraSum} \\  \cline{3-7} 
& Full T3                 & 25.28 & 5.88 & 23.00 & 2.15 & 95.91 \\
& w/o summary  & 26.40 & 6.34 & 23.67 & 2.27 & 94.86\\
& w/o qa     & 23.28 & 5.14 & 20.73 & 1.98 & 93.69\\
\cline{1-7}
\multirow{8}{*}{GPT-4-turbo} & & \multicolumn{5}{|c|}{NLQuAD} \\  \cline{3-7} 
& Full T3                 & 23.66 & 7.36 & 21.62 & 0.36 & 49.16 \\
& w/o summary  & 27.78 & 9.83 & 27.31 & 0.64 & 46.00  \\
& w/o qa     & 29.78 & 9.03 & 25.61 & 0.12 & 47.22 \\
 \cline{2-7}  
 & & \multicolumn{5}{|c|}{BBC Summary} \\  \cline{3-7} 
& Full T3                 & 36.83 & 14.14 & 33.27 & 7.69 & 69.01 \\
& w/o summary  & 37.49 & 16.46 & 35.97 & 8.72 & 63.56\\
& w/o qa     & 36.45 & 15.26 & 34.07 & 7.01 & 62.50\\
 \cline{2-7}  
 & & \multicolumn{5}{|c|}{FairyTaleQA} \\  \cline{3-7} 
& Full T3                 & 14.14 & 3.36 & 12.84 & 0.10 & 97.67\\
& w/o summary  & 16.14 & 3.66 & 14.44 & 0.19 & 97.49\\
& w/o qa     & 15.52 & 3.18 & 13.14 & 0.11 & 96.52\\
 \cline{2-7}  
\multirow{1}{*}{GPT-4-turbo}& & \multicolumn{5}{|c|}{NarraSum} \\  \cline{3-7} 
& Full T3                 & 27.27 & 7.42 & 24.45 & 3.60 & 97.44\\
& w/o summary  & 28.12 & 5.99 & 24.90 & 2.36 & 95.32\\
& w/o qa     & 27.90 & 5.96 & 24.02 & 1.98 & 94.28\\
\cline{1-7}
\multirow{16}{*}{GPT-4o} & & \multicolumn{5}{|c|}{NLQuAD} \\  \cline{3-7} 
& Full T3                 & 31.00 & 13.03 & 29.27 & 1.42 & 63.95 \\
& w/o summary  &  36.28 & 15.29 & 34.02 & 3.09 & 77.85  \\
& w/o qa     & 41.16 & 20.29 & 38.99 & 5.31 & 65.23 \\
 \cline{2-7}  
& & \multicolumn{5}{|c|}{BBC Summary} \\  \cline{3-7} 
& Full T3                 & 41.71 & 21.45 & 39.44 & 10.94 & 70.12\\
& w/o summary  & 40.17 & 19.98 & 38.35 & 9.13 & 68.56\\
& w/o qa     & 42.13 & 20.40 & 39.02 & 11.06 & 66.37\\
 \cline{2-7}  
 & & \multicolumn{5}{|c|}{FairyTaleQA} \\  \cline{3-7} 
& Full T3                 & 14.86 & 3.72 & 13.78 & 0.07 & 98.92\\
& w/o summary  & 23.80 & 9.47 & 24.14 & 0.23 & 97.35\\
& w/o qa     & 20.60 & 6.32 & 19.32 & 0.08 & 96.83\\
 \cline{2-7}  
 & & \multicolumn{5}{|c|}{NarraSum} \\  \cline{3-7} 
& Full T3                 & 28.86 & 8.58 & 25.66 & 2.55 & 99.69\\
& w/o summary  & 27.44 & 7.28 & 24.83 & 2.45 & 98.50\\
& w/o qa     & 29.03 & 7.22 & 26.04 & 2.71 & 97.42\\
 \cline{1-7} 
\multirow{16}{*}{Gemini-1.0-pro} & & \multicolumn{5}{|c|}{NLQuAD} \\  \cline{3-7} 
& Full T3                 & 28.11 & 11.05 & 26.48 & 1.25 & 71.64 \\
& w/o summary  & 25.50 & 8.99 & 23.96 & 0.42 & 66.43\\
& w/o qa     & 25.21 & 8.78 & 23.45 & 0.45 & 68.09 \\
 \cline{2-7}  
 & & \multicolumn{5}{|c|}{BBC Summary} \\  \cline{3-7} 
& Full T3                 & 33.58 & 13.92 & 31.17 & 7.22 & 55.61\\
& w/o summary  & 31.67 & 11.98 & 30.06 & 6.38 & 45.87 \\
& w/o qa     & 31.16 & 10.31 & 29.85 & 5.86 & 42.15\\
 \cline{2-7}  
 & & \multicolumn{5}{|c|}{FairyTaleQA} \\  \cline{3-7} 
& Full T3                 & 13.43 & 2.43 & 13.39 & 0.14 & 94.24 \\
& w/o summary  & 13.97 & 3.13 & 13.07 & 0.05 & 96.48\\
& w/o qa     & 13.02 & 2.12 & 12.18 & 0.01 & 95.07\\
 \cline{2-7}  
 & & \multicolumn{5}{|c|}{NarraSum} \\  \cline{3-7} 
& Full T3                 & 28.25 & 5.74 & 26.36 & 2.09 & 91.22\\
& w/o summary  & 27.63 & 5.91 & 24.59 & 1.67 & 93.70\\
& w/o qa     & 26.73 & 5.47 & 24.06 & 2.08 & 92.57\\
\cline{1-7} 

\multirow{10}{*}{Gemini-1.5-pro} & & \multicolumn{5}{|c|}{NLQuAD} \\  \cline{3-7} 
& Full T3                 & 24.19 & 7.97 & 22.44 & 0.47 & 65.09 \\
& w/o summary  & 25.51 & 8.00 & 24.48 & 0.35 & 68.40 \\
& w/o qa     & 26.98 & 8.08 & 24.93 & 0.42 & 67.36 \\
 \cline{2-7}  
& & \multicolumn{5}{|c|}{BBC Summary} \\  \cline{3-7} 
& Full T3                 & 35.10 & 12.58 & 31.69 & 6.47 & 60.49 \\
& w/o summary  & 37.16 & 14.76 & 33.21 & 7.03 & 68.21\\
& w/o qa     & 37.14 & 15.77 & 34.70 & 9.02 & 65.94\\
 \cline{2-7}  
& & \multicolumn{5}{|c|}{FairyTaleQA} \\  \cline{3-7} 
& Full T3                 & 17.56 & 4.34 & 15.46 & 0.13 & 98.53 \\
& w/o summary  & 14.63 & 3.24 & 13.94 & 0.04 & 97.74 \\
& w/o qa     & 14.50 & 3.20 & 13.50 & 0.02 & 97.20 \\
 \cline{2-7}  
\multirow{1}{*}{Gemini-1.5-pro} && \multicolumn{5}{|c|}{NarraSum} \\  \cline{3-7} 
& Full T3                 & 26.95 & 5.39 & 25.32 & 1.99 & 93.57\\
& w/o summary  & 25.23 & 5.97 & 24.37 & 2.19 & 93.04\\
& w/o qa     & 26.36 & 6.00 & 26.04 & 2.00 & 92.61\\
\cline{1-7}
\multirow{16}{*}{Claude-3.0-sonnet} & & \multicolumn{5}{|c|}{NLQuAD} \\  \cline{3-7} 
& Full T3                 & 33.72 & 13.27 & 31.22 & 2.37 & 61.69\\
& w/o summary  & 35.42 & 14.30 & 32.81 & 2.51 & 68.94 \\
& w/o qa     & 33.37 & 12.20 & 30.82 & 3.17 & 63.14 \\
 \cline{2-7}  
 & & \multicolumn{5}{|c|}{BBC Summary} \\  \cline{3-7} 
& Full T3                 & 40.90 & 18.08 & 37.39 & 10.74 & 50.43\\
& w/o summary  & 40.94 & 18.85 & 36.98 & 10.06 & 52.32\\
& w/o qa     & 40.27 & 17.63 & 36.93 & 10.09 & 48.74\\
 \cline{2-7}  
 & & \multicolumn{5}{|c|}{FairyTaleQA} \\  \cline{3-7} 
& Full T3                 & 18.91 & 4.90 & 17.35 & 0.33 & 98.92\\
& w/o summary  & 18.24 & 4.35 & 16.83 & 0.24 & 99.36\\
& w/o qa     & 15.21 & 3.82 & 14.58 & 0.08 & 97.54\\
 \cline{2-7}  
 & & \multicolumn{5}{|c|}{NarraSum} \\  \cline{3-7} 
& Full T3                 & 25.87 & 5.47 & 22.51 & 2.05 & 95.61\\
& w/o summary  & 28.34 & 6.57 & 26.05 & 2.47 & 97.59 \\
& w/o qa     & 27.52 & 6.09 & 25.21 & 2.02 & 94.83 \\
\cline{1-7}
\multirow{16}{*}{Claude-3.5-sonnet} & & \multicolumn{5}{|c|}{NLQuAD} \\  \cline{3-7} 
& Full T3                 & 40.17 & 18.74 & 37.92 & 5.00 & 82.76 \\
& w/o summary  & 36.79 & 16.71 & 34.45 & 4.09 & 75.03 \\
& w/o qa     & 38.08 & 15.83 & 35.80 & 4.02 & 72.94\\
 \cline{2-7}  
 & & \multicolumn{5}{|c|}{BBC Summary} \\  \cline{3-7} 
& Full T3                 & 43.54 & 21.13 & 40.36 & 13.00 & 65.83  \\
& w/o summary  & 40.26 & 16.83 & 35.99 & 13.31 & 58.32 \\
& w/o qa     & 41.55 & 20.72 & 40.05 & 31.90 & 58.57 \\
 \cline{2-7}  
 & & \multicolumn{5}{|c|}{FairyTaleQA} \\  \cline{3-7} 
& Full T3                 & 20.96 & 6.07 & 19.52 & 0.47 & 99.97\\
& w/o summary  & 19.94 & 5.84 & 19.41 & 0.53 & 98.53\\
& w/o qa     & 19.86 & 5.36 & 18.18 & 0.31 & 97.98\\
 \cline{2-7}  
 & & \multicolumn{5}{|c|}{NarraSum} \\  \cline{3-7} 
& Full T3                 & 29.78 & 7.79 & 27.23 & 3.23 & 98.14\\
& w/o summary  & 29.63 & 7.60 & 27.86 & 2.96 & 96.43 \\
& w/o qa     & 29.38 & 7.43 & 26.53 & 2.41 & 97.61\\
\hline

\caption{Ablation study results on BBC summary, NarraSum, NLQuAD and FairytaleQA}
\label{tab:Table3}
\end{longtable}

\subsection{Significance Analysis}
\label{significance}
In this analysis, we used the \textbf{t-test} to evaluate the significance of the observed results. The null hypothesis \( H_0 \) assumes that there is no significant difference between the two algorithms compared in our experiment. The test statistic \( t \) is calculated as follows:

\begin{equation}
    t = \frac{\bar{x}_1 - \bar{x}_2}{\sqrt{\frac{s_1^2}{n_1} + \frac{s_2^2}{n_2}}}
    \label{eq:t-test}
\end{equation}
where\( \bar{x}_1, \bar{x}_2 \) are the sample means of two groups,  \( s_1, s_2 \) are the sample standard deviations,  
 \( n_1, n_2 \) are the sample sizes.

We calculate the sample means \( \bar{x}_1, \bar{x}_2 \) and compute the sample standard deviations \( s_1, s_2 \).  
Then we plug values into Equation~\eqref{eq:t-test} to determine the test statistic \( t \) and compare \( t \)-value with the critical value from the \( t \)-distribution at a given confidence level \( \alpha = 0.05 \).

The p-value for each metric result between our direct summarization and summary with T3 are listed in Table~\ref{tab:significance_values}. Here we sign the model with * if the p-values for the metric are less than 0.05. 

\begin{table*}[ht!]
\centering

\begin{tabular}{lccccc}
\toprule
Models & ROUGE-1 & ROUGE-2 & ROUGE-L & BLEU  &  Factscore\\ \midrule

& \multicolumn{5}{c}{BBC summary} \\ \cmidrule(lr){2-6} 
% &  w/o  \hspace{0.2cm}| \hspace{0.3cm}T3 &  w/o  \hspace{0.2cm}| \hspace{0.3cm}T3 &  w/o  \hspace{0.2cm}| \hspace{0.3cm}T3  &  w/o  \hspace{0.2cm}| \hspace{0.3cm}T3  &  w/o  \hspace{0.2cm}| \hspace{0.3cm}T3
%                 \\ \cmidrule(lr){2-6} 
 
GPT-4o*                 & 0** & 0** & 0**   &0.0007    &  0.0004   \\

GPT-4-turbo &  0.0431 & 0.1714 & 0.0367 & 0.0836   & 0.0227 \\
GPT-3.5-turbo* & 0.0124  & 0.0388 & 0.0080 &0.0082 & 0.0017 \\
Gemini-1.5-pro* & 0.0160 & 0.0089 & 0.0011 & 0.0157 & 0.0160 \\
Gemini-1.0-pro* & 0.0101 & 0.0191 & 0.0103 & 0.0495  & 0.0091  \\
 
Claude-3.5-sonnet* & 0.0042 & 0.0003 & 0.0043 &0.0001 &  0.0003 \\
Claude-3.0-sonnet & 0.0795 & 0.1108 & 0.0292 & 0.0268 &  0.5309 \\

 \midrule
& \multicolumn{5}{c}{NarraSum} \\ \cmidrule(lr){2-6} 
% &  w/o  \hspace{0.2cm}| \hspace{0.3cm}T3 &  w/o  \hspace{0.2cm}| \hspace{0.3cm}T3 &  w/o  \hspace{0.2cm}| \hspace{0.3cm}T3  &  w/o  \hspace{0.2cm}| \hspace{0.3cm}T3  &  w/o  \hspace{0.2cm}| \hspace{0.3cm}T3
%                 \\ \cmidrule(lr){2-6} 
GPT-4o*                  &  0** & 0.0009 & 0**    &  0.0407  & 0.0003   \\

GPT-4-turbo* & 0.0077 & 0.0114 & 0.0177 & 0.0418  & 0.3652 \\
GPT-3.5-turbo* &0.0499 &0.0431 & 0.0478 & 0.0456 &  0.0203 \\

Gemini-1.5-pro* & 0.0332 & 0.0074 & 0.0196 & 0.0482 & 0.0459 \\
Gemini-1.0-pro & 0.0198 & 0.0312 & 0.0422 & 0.0696  & 0.0086 \\

Claude-3.5-sonnet & 0.0182 & 0.0417 & 0.0152 & 0.1526 & 0.1040 \\
Claude-3.0-sonnet* & 0.0483 & 0.0099 & 0.0523 & 0.0154 &  0.0691 \\

 \midrule

& \multicolumn{5}{c}{NLQuAD} \\ \cmidrule(lr){2-6} 
% &  w/o  \hspace{0.2cm}| \hspace{0.3cm}T3 &  w/o  \hspace{0.2cm}| \hspace{0.3cm}T3 &  w/o  \hspace{0.2cm}| \hspace{0.3cm}T3  &  w/o  \hspace{0.2cm}| \hspace{0.3cm}T3  &  w/o  \hspace{0.2cm}| \hspace{0.3cm}T3 
%                 \\ \cmidrule(lr){2-6} 
 
GPT-4o* & 0.0086 & 0.0084 & 0.0072 & 0.0087 & 0.0420  \\
GPT-4-turbo* & 0.0189 & 0.0417 & 0.0162 & 0.0488 & 0.0340  \\

GPT-3.5-turbo* & 0.0299  & 0.0441 & 0.0376 &0.0276 & 0.0457 \\
Gemini-1.5-pro* & 0.0436 & 0.0468 & 0.0282 & 0.0261 & 0.0014 \\

Gemini-1.0-pro* & 0.0479  & 0.0420 & 0.0289 & 0.0181  & 0.0498 \\
Claude-3.5-sonnet* & 0.0061  & 0.0003 & 0.0034 & 0.0002  &  0.0212 \\

Claude-3.0-sonnet* & 0.0095 & 0.0403 & 0.0427 & 0.0193  & 0.0089 \\

 \midrule
& \multicolumn{5}{c}{FairytaleQA} \\ \cmidrule(lr){2-6} 
               % &  w/o  \hspace{0.2cm}| \hspace{0.3cm}T3 &  w/o  \hspace{0.2cm}| \hspace{0.3cm}T3 &  w/o  \hspace{0.2cm}| \hspace{0.3cm}T3  &  w/o  \hspace{0.2cm}| \hspace{0.3cm}T3  &  w/o  \hspace{0.2cm}| \hspace{0.3cm}T3\\ \cmidrule(lr){2-6} 
 GPT-4o* & 0.0083 & 0.0180 & 0.0100 & 0.0194 & 0.0124  \\
GPT-4-turbo* & 0.0073 & 0.0375 & 0.0185 & 0.0367  & 0.0356 \\
GPT-3.5-turbo & 0.0409 & 0.0524 & 0.0895 & 0.0451 &  0.1184 \\
Gemini-1.5-pro* & 0.0473 & 0.0296 & 0.0188 & 0.0298 &  0.0365 \\
Gemini-1.0-pro* & 0.0168 & 0.0151 & 0.0169 & 0.0355 & 0.0496 \\
Claude-3.5-sonnet* & 0.0204 & 0.0031 & 0.0239 & 0.0423 & 0.0225 \\
Claude-3.0-sonnet* & 0.0340 & 0.0176 & 0.0338& 0.0147 &  0.0092 \\

\bottomrule
\end{tabular}

\caption{Experimental Results of the p-value on BBC summary, NarraSum, NLQuAD and FairytaleQA datasets between the result of each LLM without and with T3. 0** means the p-value is less than $10^{-7}$}
\label{tab:significance_values}
\end{table*}

\end{document}